\documentclass{article}
\usepackage[final]{neurips_2025}

\usepackage[utf8]{inputenc} 
\usepackage[T1]{fontenc}    
\usepackage{hyperref}       
\usepackage{url}            
\usepackage{booktabs}       
\usepackage{amsfonts}       
\usepackage{nicefrac}       
\usepackage{microtype}      
\usepackage{xcolor}         

\usepackage{amssymb}
\usepackage{graphicx}
\usepackage{mathtools}
\usepackage{algorithm}
\usepackage{algpseudocode}
\usepackage{multirow} 
\usepackage{color, colortbl, xcolor}
\usepackage{amsmath}
\usepackage{amssymb}
\usepackage{booktabs}
\usepackage{natbib}
\usepackage{subcaption}
\usepackage{enumitem}
\usepackage{wrapfig}
\usepackage{comment}

\newcommand{\method}{Hyper-GoalNet}

\title{
{\it Hyper-GoalNet}: Goal-Conditioned Manipulation Policy Learning with HyperNetworks
}

\author{{
Pei Zhou$^{1}$\ \ ~ 
Wanting Yao$^{1,2}$\thanks{Work done as an intern at HKU.}\ \ ~ 
Qian Luo$^{1}$\ \ ~
Xunzhe Zhou$^{1}$\ \ ~
Yanchao Yang$^{1}$}
\\
\\
{
$^{1}$InfoBodied AI Lab, The University of Hong Kong}\ \ \ 
{$^2$University of Pennsylvania}\\
{\texttt{\{pezhou,qianluo1,xunzhe\_zhou\}@connect.hku.hk}}\\
{\texttt{wtyao@seas.upenn.edu},
\texttt{yanchaoy@hku.hk}}
}

\begin{document}
\maketitle

\begin{abstract}
Goal-conditioned policy learning for robotic manipulation presents significant challenges in maintaining performance across diverse objectives and environments. We introduce {\it Hyper-GoalNet}, a framework that generates task-specific policy network parameters from goal specifications using hypernetworks. Unlike conventional methods that simply condition fixed networks on goal-state pairs, our approach separates goal interpretation from state processing -- the former determines network parameters while the latter applies these parameters to current observations. To enhance representation quality for effective policy generation, we implement two complementary constraints on the latent space: (1) a forward dynamics model that promotes state transition predictability, and (2) a distance-based constraint ensuring monotonic progression toward goal states. We evaluate our method on a comprehensive suite of manipulation tasks with varying environmental randomization. 
Results demonstrate significant performance improvements over state-of-the-art methods, particularly in high-variability conditions. Real-world robotic experiments further validate our method's robustness to sensor noise and physical uncertainties. 
Code is available at: 
\enspace\texttt{\href{https://github.com/wantingyao/hyper-goalnet}{\color{blue} https://github.com/wantingyao/hyper-goalnet}}.
\end{abstract}

\section{Introduction}
\label{sec:intro}

Goal-conditioned policy learning 
enables embodied agents to adjust their actions 
based on current state observations 
and specified goals~\cite{chen2024igor,liu2022goal,seita2021learning}. By integrating goal information into decision making, 
agents leverage knowledge across various tasks, enhancing adaptability~\cite{buclosed,ding2019goal,nasiriany2019planning} in hierarchical reinforcement learning and complex imitation learning~\cite{bacon2017option, florensa2017stochastic,sutton2018reinforcement}. 

Conventional approaches typically concatenate goal observations with current states as input to a fixed-parameter network~\cite{cui2022play,yu2020meta,wang2023mimicplay}. This design creates a fundamental limitation: the network must process all possible goal-current state combinations using the same fixed weights, conflating ``what'' to process (current state) with ``how'' to process it (goal-dependent strategy). 
Consequently, these architectures often struggle with generalization to novel goals and complex manipulation tasks that require different processing strategies depending on the goal specification.

We {\it aim to} rethink this relationship by treating goals {\it not} as additional input features but as specifications that determine {\it how} current observations should be processed. 
Hypernetworks -- neural networks that generate weights for another network -- offer a natural implementation of this perspective. 
By explicitly modeling goals as determinants of policy parameters rather than as inputs, 
hypernetworks effectively disentangle task-dependent processing (defined by goals) from state-dependent processing (applied to current observations)~\cite{galanti2020modularity,schug2024attention}. 
This approach better aligns with biological goal-directed behavior, where prefrontal regions interpret task goals and dynamically modulate processing in sensorimotor circuits accordingly~\cite{miller2001integrative,ramnani2001cerebellum}.

\begin{figure}[t]
    \centering
    \includegraphics[width=0.92\linewidth]{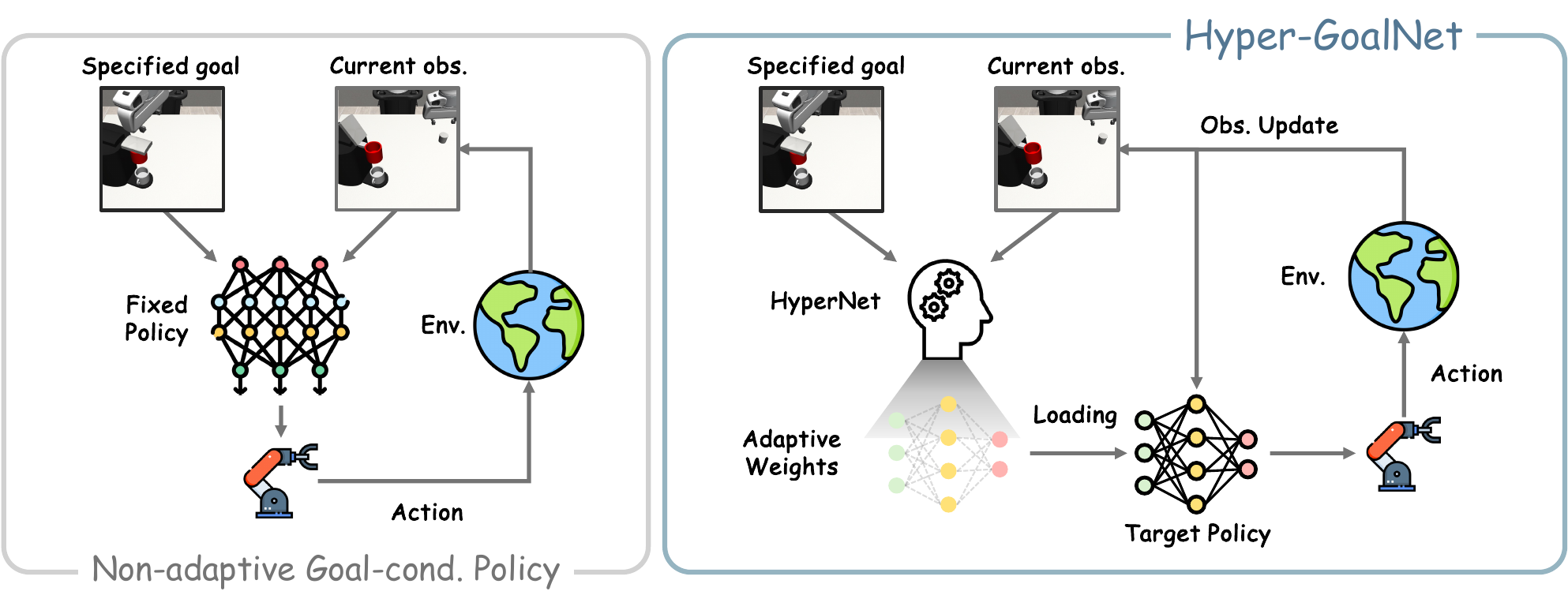}
    \caption{\textbf{The proposed 
    Goal-Conditioned Policy Generation 
    framework (\method)
    and conventional goal-conditioned policies.
    } 
    Existing methods typically employ a fixed-parameter policy network 
    that processes concatenated current observations and goal states, 
    treating goals mostly as additional inputs. 
    In {\it contrast,} 
    our approach formulates policy learning 
    as an adaptive generation task, 
    where the goal image determines the parameters of the policy network itself -- 
    transforming goals from inputs into specifications that define {\it how} current observations should be processed.
    This allows for more effective handling of diverse goals and complex manipulation tasks.}
    \label{fig:teaser}
    \vspace{-5mm}
\end{figure}

To address these challenges, 
we present {\bf \method}, 
a hypernetwork-based framework for robotic manipulation illustrated in Fig.~\ref{fig:teaser}. 
Our approach employs a hypernetwork to dynamically generate target policy parameters conditioned on specified goals. 
When loaded into the policy network, these parameters enable the processing of current observations without requiring further access to goal information. 
This architecture creates a clear separation of concerns: the hypernetwork interprets what the goal means for processing strategy, while the generated policy focuses exclusively on transforming current observations into appropriate actions. 
A key advantage of this goal-aware design is that it enables the system to autonomously detect task completion during execution.
By training the hypernetwork to model 
the conditional distribution of effective policy weights 
given goal specifications, 
we obtain a system that can adapt its processing strategy to diverse manipulation tasks.

Our technical contributions 
center on effectively applying hypernetworks for parameter-adaptive goal-conditioned policies learning. 
{\it First,} we adapt optimization-inspired hypernetwork architectures for generating policy parameters conditioned on goal specifications, 
creating a framework that dynamically determines how current observations should be processed. 
{\it Second,} we introduce an effective 
latent space shaping technique that imposes two critical properties: (1) predictability of future states through a learned dynamics model, and (2) preservation of physical relationships through distance constraints that ensure monotonic progression toward goals. 
These properties create an ideal representation space for our hypernetwork, 
providing clear signals about how policy parameters should change as states approach goals.

Our extensive experiments 
across multiple manipulation tasks show that 
\method{} significantly outperforms state-of-the-art methods, achieving higher success rates on complex contact-rich manipulations. 
Notably, while conventional approaches fail almost completely in high-variability environments, 
our method maintains robust performance. 
Ablation studies confirm the critical importance of our proposed components in the parameter-adaptive goal-conditioned policy leaning framework. 
Finally, real-robot experiments demonstrate that our parameter-adaptive approach succeeds in physical environments where conventional methods struggle with sensor noise and environmental variations. 
These results confirm that explicitly modeling goals as determinants of processing strategy rather than as additional inputs creates a more effective and robust framework for goal-conditioned manipulation.

\section{Related Work}

\textbf{Goal-conditioned policy.} 
Goal-conditioned policy learning has attracted significant attention for developing versatile and generalizable agents~\cite{kaelbling1996reinforcement,vuong2023open,team2024octo,zhouautocgp,kim2024openvla}. Traditional methods augment state spaces with goal information and train policies that condition on these augmented states~\cite{du2023learning,schaul2015universal,ma2022offline,xu2025diffusion}. Hindsight Experience Replay (HER)~\cite{andrychowicz2017hindsight} exemplifies this approach by allowing agents to learn from failures by reinterpreting unsuccessful outcomes as alternative goals. However, these methods typically suffer from increased complexity and require extensive tuning to manage the parameters associated with goal conditioning~\cite{pong2018temporal,nair2018visual,sridhar2024nomad}. Our approach distinguishes itself by utilizing hypernetworks to dynamically generate policy weights, thereby reducing parameters in the policy network while enhancing scalability without extensive retraining~\cite{ha2018world, santos2019learning}.

In the realm of imitation learning for goal-conditioned policies, several frameworks have demonstrated promising results by learning from pre-collected demonstrations~\cite{lynch2020learning, cui2022play, wang2023mimicplay}. Recent goal-conditioned behavior cloning approaches such as C-BeT~\cite{cui2022play} and MimicPlay~\cite{wang2023mimicplay} have advanced long-horizon manipulation tasks. However, these methods typically require sequences of achievable goal images, which are challenging to obtain in practice. Moreover, while performing well in basic pick-and-place scenarios, they often struggle with contact-rich tasks that demand precise environmental awareness~\cite{mandlekar2023mimicgen}. Our method overcomes these limitations through effective latent space shaping, requiring only a single goal image while maintaining robustness across diverse manipulation scenarios.

\textbf{Hypernetworks and Cognitive science insights for goal-directed behavior.} Our work draws inspiration from cognitive science research on human goal-directed behavior, where meta-cognitive strategies and higher-level planning mechanisms enable adaptive actions~\cite{botvinick2009hierarchically, o2006making, pezzulo2016navigating}. Studies show that humans efficiently manage cognitive resources and flexibly adapt to different goals through higher-level representations~\cite{frank2004carrot, dayan2008decision, daw2005uncertainty}. Current policy learning methods incorporating cognitive principles often focus on imitation learning to mimic human strategies~\cite{argall2009survey, finn2016unsupervised, duan2016benchmarking}, but can be limited by demonstration quality and diversity~\cite{pomerleau1988alvinn, ross2011reduction, ho2016generative}. 
Hypernetworks have been explored in robotic control~\cite{hegde2024hyperppo,yu2023hyper,auddy2023scalable}, though primarily within reward-driven reinforcement learning (RL) settings~\cite{huang2021continual,beck2023hypernetworks}. This fundamental difference in training paradigms, RL versus our reward-free behavior cloning (BC), means their end-to-end algorithms are not directly adaptable. We clarify, however, that their core hypernetwork architectures can be decoupled from the RL framework.
By embedding cognitive insights into our hypernetwork architecture, we emulate human-like flexible adaptation~\cite{botvinick2009hierarchically, o2006making, pezzulo2016navigating}. \method's capacity to generate goal-specific policy parameters without extensive retraining addresses practical challenges of goal-conditioned learning~\cite{ha2018world} while mirroring key cognitive mechanisms, offering a biologically plausible framework for adaptable policy generation.
\section{Method}
\label{sec:method}

Let
\( \mathcal{D} = \{ \tau_i \}_{i=1}^{M} \)
be a dataset 
consisting of \( M \) 
robotic manipulation demonstrations,
where
each trajectory comprises a sequence of observation-action pairs, 
i.e.,
\( \tau_i = \{ (o^i_j, a^i_j) \}_{j=1}^{N_i} \), 
with \( a_j \) denoting a continuous-valued action 
and \( o_j \) representing a tuple containing high-dimensional state observations. 
{\it Specifically,} 
\( o_j \) includes an RGB image \( I_j \) 
captured by a single front-view camera, 
as well as the proprioceptive information \( s_j \) of the embodied agent.
Given this formulation, our objective is to develop 
a {\it generalizable goal-conditioned policy learning framework} 
that enables efficient adaptation to diverse manipulation tasks.

We propose a shift from conventional 
goal-conditioned policies, 
which typically use fixed parameters while processing both current and goal images. 
Our key insight is that the goal image inherently specifies {\it how} the current image should be processed to generate appropriate actions. 
Therefore, we argue that the policy parameters themselves -- 
which determine the processing mechanism -- 
should adapt based on different goal specifications. 
To realize this insight, 
we leverage hypernetworks to dynamically generate task-specific policy parameters conditioned on goal images, rather than directly conditioning a fixed policy network on both current and goal observations. 
This approach creates a more flexible and efficient framework where the processing of current states is explicitly tailored to the specified goals.

\begin{figure*}[!t]
  \centering
   \includegraphics[width=\linewidth]{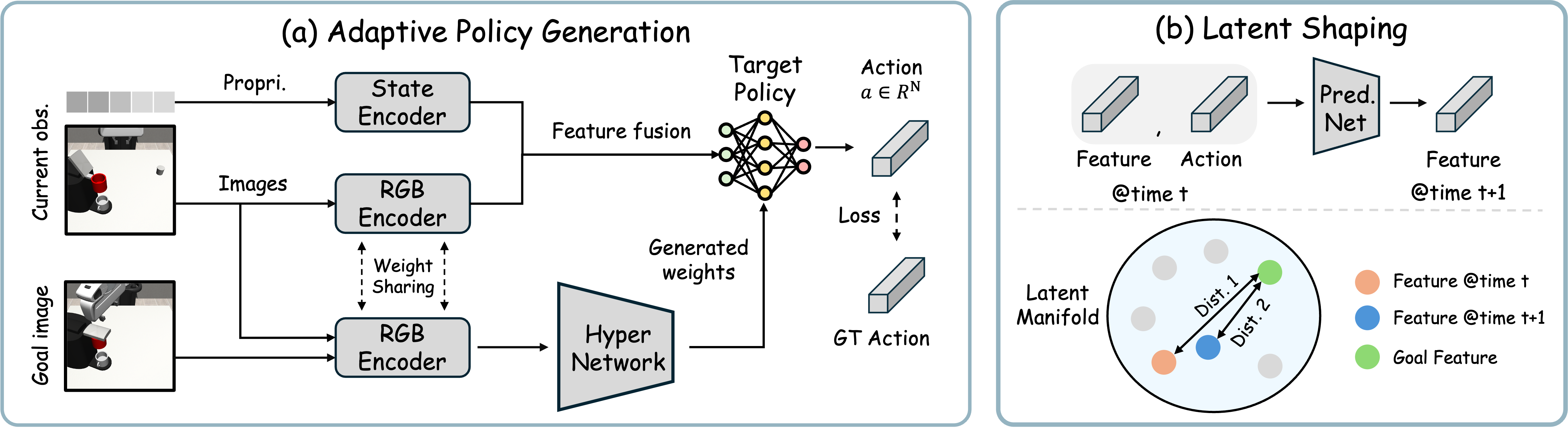}
   \caption{\textbf{
   An overview of the proposed \method{} framework.} 
   (a) \textit{Adaptive Policy Generation:} Unlike conventional approaches with fixed parameters, our hypernetwork dynamically generates task-specific policy parameters conditioned on goal images. This creates a parameter-adaptive target policy that processes current observations (RGB images and proprioception) through a multimodal encoder to produce actions tailored to specific goals.
   (b) \textit{Latent Shaping:} Our approach enhances performance by explicitly structuring the latent space in two ways: a predictive network models state transitions to improve temporal dynamics, while geometric constraints ensure distances to goal states monotonically decrease during successful trajectories (detailed in Sec.~\ref{subsec:gr_policy}).}
   \label{fig:overview_method}
\end{figure*}

The full pipeline is illustrated in Fig.~\ref{fig:overview_method}. 
{\it Next,} 
we elaborate on the key components of our approach. 
In Sec.~\ref{subsec:hp_gen}, 
we describe how we adapt hypernetwork architectures 
to effectively generate varying goal-reaching policies. 
In Sec.~\ref{subsec:gr_policy}, 
we introduce an effective 
latent space shaping techniques 
that significantly enhance the performance of goal-conditioned policy generation 
by enforcing meaningful geometric structure in the representation space. 
{\it Finally,}
in Sec.~\ref{subsec:GCP},
we present the test‑time inference pipeline that integrates our trained model (\method) to accomplish diverse manipulation tasks, 
highlighting the practical advantages of our parameter-adaptive approach.

\subsection{Goal-Conditioned Hypernetworks}
\label{subsec:hp_gen}

We formulate goal-conditioned policy learning as a parameter generation task rather than a direct conditioning problem. 
This formulation reframes the challenge from ``what action to take given current and goal observations'' to ``what processing parameters to use given the goal.''

More formally, given a current observation \(o_c\) and a desired goal observation \(o_g\), we model the conditional distribution over target policy weights that will transform the scene into the goal configuration. With the set of robotic manipulation demonstrations $\mathcal{D}$, we learn the distribution $\mathcal{H}(\theta \mid o_c, o_g)$:
\begin{align}
\mathcal{H}(\theta \mid o_c, o_g) \coloneqq \mathbb{P}_{\mathcal{D}}\bigl(\theta \mid o_c = I_{t},\,o_g = I_{t'}\bigr),\ \text{where }I_{t}, I_{t'}\in\tau_i\in\mathcal{D}\text{ and }t'>t.
\end{align}
For practical implementation, we focus on a single goal state rather than a sequence of goals, and use RGB image observations to condition the hypernetwork. Since our primary objective is to investigate the efficacy of goal-conditioned policy generation for manipulation tasks rather than developing a full probabilistic model, we approximate this as a deterministic mapping:
\begin{equation}
    \mathcal{H} : \mathcal{O} \times \mathcal{O} \to \Theta,
\end{equation}
where \( \mathcal{O} \) denotes the observation space and \( \Theta \) represents the space of target policy parameters. With a current-goal observation pair \( (o_c, o_g) \), the hypernetwork \(\mathcal{H}\) produces a policy that guides the transition from current state \( o_c \) to goal state \( o_g \) through action execution.

\textbf{Hypernetwork architecture.} 
To implement our parameter-adaptive approach, 
we adopt a hypernetwork architecture that efficiently generates policy parameters for achieving specified goals. 
The architecture must be capable of 
capturing the complex relationships between current states, goal states, and the required actions to bridge them.

We leverage an optimization-inspired architecture following~\cite{renhypogen}, 
which provides beneficial inductive bias for our parameter generation task. 
This approach mimics iterative optimization by refining policy parameters through multiple feed-forward steps:
\begin{equation}
    \theta^K = \mathcal{H}(o_c, o_g),
\end{equation}
where $\theta^K$ represents the final policy parameters after $K$ refinement iterations, with each update computed as:
\begin{equation}
\theta^k = \theta^{k-1} + \lambda^k(\theta^{k-1}, \alpha)\,\psi^k(\theta^{k-1}, \alpha),\ \alpha = \phi(o_c, o_g).
\label{eq:iter-update}
\end{equation}
Here, neural modules \(\lambda^k\) and \(\psi^k\) serve as learned analogs to step sizes and gradients in optimization, operating on embeddings \(\phi(o_c, o_g)\) of the current and goal observations. This mechanism enhances the hypernetwork's ability to generate effective task-specific policy parameters and improves generalization to new goal specifications. Once we obtain the goal-conditioned policy weights \(\theta\), we can process the current observation through the generated policy to predict appropriate actions.

\textbf{Hypernetwork Training.} We train our hypernetwork $\mathcal{H}: \mathcal{O}\times\mathcal{O} \to \Theta$ to generate parameters for the target visuomotor policy $\pi(\cdot; \theta)$ using behavior cloning (BC) on demonstration data. The generated policy takes the current observation $o_t$ (comprising image $I_t$ and proprioception $s_t$) and outputs actions for execution. 

To enhance robustness, we utilize a sequence of $L$ consecutive observations as input to the policy, capturing temporal dependencies under a non-Markovian assumption. The training objective minimizes the BC loss between demonstrated actions \(a^i_t\) and predicted actions \(\hat a^i_t\):
\begin{equation}
\mathcal{L}_{\mathrm{policy}}
= \sum_{i=1}^M \sum_{\substack{1 \le t < t' \le N_i}}
\ell\bigl(a^i_t,\;\hat a^i_t\bigr),
\quad
\hat a^i_t = \pi\bigl(o^i_{t-L:t};\,\mathcal{H}(o^i_t, o^i_{t'})\bigr).
\end{equation}
Here, \(\ell\) represents the Mean Squared Error (MSE) loss. The end-to-end training procedure works as follows: for a given current observation \(o^i_t\) and a goal image \(o^i_{t'}\), the hypernetwork \(\mathcal{H}\) generates the weights for the policy network \(\pi\). This goal-conditioned policy then processes the observation sequence \(o^i_{t-L:t}\) to predict the action \(\hat{a}^i_t\). The resulting loss is backpropagated through both the policy network and the hypernetwork. We restrict goal representations to image inputs since proprioceptive goal states may not always be available in practical applications. This formulation allows the hypernetwork to learn how to generate goal-specific processing mechanisms (target policy parameters) from visual goals, embodying our key insight that goal images determine {\it how} current observations should be processed.

\subsection{Latent Space Shaping}
\label{subsec:gr_policy}

A critical insight in our approach is that the effectiveness of parameter-adaptive policies depends significantly on the quality of the representation space in which observations are embedded. 
While our hypernetwork can directly generate policy parameters from raw observations, 
we find that explicitly shaping the latent representation space substantially enhances performance. 
Given the high dimensionality and redundant information in RGB images, we employ an image encoder $\mathcal{E}$ to extract task-relevant features and compress them into low-dimensional latents $z = \mathcal{E}(I)$.

We identify two fundamental properties that, 
when enforced in the latent space, particularly benefit our parameter-adaptive approach: {\it predictability} and {\it physical structure preservation}. The first property ensures the latent space facilitates modeling of state transitions, 
making the hypernetwork's task of generating appropriate policy parameters more tractable. 
The second property ensures that the geometric relationships in the latent space meaningfully reflect physical relationships between states, enabling the generated policies to exploit these structured representations.

{\bf Enhancing predictability through dynamic modeling.} 
To improve the predictability of latent representations, we introduce a dynamics model that forecasts future states in the latent space. By training this model to predict state transitions while simultaneously shaping the encoder $\mathcal{E}$ through backpropagation, we create a latent space where sequential relationships are explicitly captured. This significantly benefits our hypernetwork, as it needs to generate policy parameters that leverage these sequential relationships to guide transitions from current to goal states.

Formally, consider a discrete-time dynamical system with state representation $z_t \in \mathcal{Z}$ and control input $a_t \in \mathcal{A}$. The forward dynamic model is:
\begin{equation}
    \hat{z}_{t+1} \sim p_{\Phi}(z_{t+1} \mid z_t, a_t),
\end{equation}
where $z_t = \mathcal{E}(I_t)$ represents the latent encoding at time $t$, $a_t$ is the executed action, and $p_{\Phi}$ is the transition dynamics parameterized by $\Phi$. For practical implementation, we approximate this with a deterministic model $\Phi: \mathcal{Z} \times \mathcal{A} \rightarrow \mathcal{Z}$. 
The corresponding learning objective minimizes 
the prediction loss:
\begin{equation}
    \mathcal{L}_{\text{pred}} 
    = \mathbb{E}_{\tau \sim \mathcal{D}} 
    \left[ 
    \ell(\Phi(z_t, a_{t}), z_{t+1}) \right],
\end{equation}
where $\ell$ is a distance metric in the latent space. When $\Phi$ is well-trained, we further finetune $\mathcal{E}$ through $\Phi$ to shape representations that capture both current state and potential transition information.

{\bf Preserving physical structure through distance constraints.} 
For our parameter-adaptive approach to be effective, the latent space must preserve the physical structure of the task, particularly the progression towards goals. We formalize this as a requirement that the distance between the current state and goal state should monotonically decrease along goal-reaching trajectories. This property is especially valuable for our hypernetwork, as it provides a clear signal about how policy parameters should change as states approach goals.

Specifically, for any goal-reaching trajectory $\tau_i \in \mathcal{D}$, where $\tau_i = \{ (o^i_j, a^i_j) \}_{j=1}^{N_i}$, we enforce:
\begin{equation}
d_\mathcal{E}(o^i_j, o^i_{j'}) \geq 
d_\mathcal{E}(o^i_{j+1}, o^i_{j'}), 
\quad \forall j < j',
\end{equation}
where $d_\mathcal{E}$ denotes a distance metric in the latent space. To explicitly model this behavior, we propose the following loss function:
\begin{equation}
    \mathcal{L}_{\mathrm{dist}}
= \mathbb{E}_{\tau\sim\mathcal{D}}
\sum_{j}\max\bigl(0,\;\beta + d(z_{j+1},z_g) - d(z_j,z_g)\bigr),
\end{equation}
where $d(z_1, z_2) = \|z_1 - z_2\|_2$ represents the Euclidean distance between latent features, $z_j = \mathcal{E}(I_j)$ and $z_g = \mathcal{E}(I_g)$ denote the image and goal image latents, respectively. The margin parameter $\beta \geq 0$ enforces a minimum decrease in distance between consecutive states and the goal. Empirically, setting $\beta=0$ suffices to induce the desired monotonic progression.

This shaped latent space creates an ideal foundation for our parameter-adaptive approach, as it encodes both the predictive dynamics and geometric structure needed for the hypernetwork to effectively generate goal-tailored policy parameters. Figure~\ref{fig:effect_shaping} illustrates how our latent space shaping approach compares to alternative methods, showing the enhanced structure that benefits our goal-conditioned policy generation.

\subsection{\method{} for Manipulation}
\label{subsec:GCP}

Having established our parameter-adaptive architecture and latent space shaping techniques, we now present our complete framework, \method{}, which synthesizes these components for effective goal-conditioned manipulation. The {\bf overall training objective} combines our policy generation loss with the latent space shaping terms:
\begin{equation}
\mathcal{L}_{\text{\method}}
= \mathcal{L}_{\mathrm{policy}}
  + \lambda_{\mathrm{pred}}\,\mathcal{L}_{\mathrm{pred}}
  + \lambda_{\mathrm{dist}}\,\mathcal{L}_{\mathrm{dist}},
\end{equation}
where $\lambda_{\text{pred}}$ and $\lambda_{\text{dist}}$ are weight coefficients balancing the contributions of predictability and structural constraints. The framework is trained end-to-end using gradient descent, allowing all components to co-adapt for optimal performance.

During {\bf inference for task completion}, \method{} generates goal-specific policy parameters by feeding the concatenated latent features $[z_g, z_t]$ into the hypernetwork $\mathcal{H}$, where $z_g = \mathcal{E}(I_g)$ and $z_t = \mathcal{E}(I_t)$ are the latent representations of the goal and current observations. The generated goal-specific parameters $\theta = \mathcal{H}(z_t, z_g)$ are then loaded into the target policy $\pi(\cdot;\theta)$, which processes the current observation sequence to produce actions that guide the agent toward the goal.

\method{} offers two principal advantages over conventional goal-conditioned policies:

{\bf 1) Parameter-adaptive policy generation:} By dynamically synthesizing policy parameters based on goal specifications, 
our approach effectively 
changes how goal-conditioned policies operate. 
Rather than relying on a fixed network with static parameters to handle all possible goals, 
\method{} generates compact and efficient 
processing pathways that are tailored to specific goals. 

{\bf 2) Natural goal completion detection:} Our shaped latent space provides an elegant solution to the challenging problem of goal completion detection. The distance metric $d_{\mathcal{E}}(o_t, o_g)$ in the latent space serves as a natural criterion for determining when a goal has been achieved, enabling autonomous goal transitions without external supervision. 

The {\bf test-time task evaluation} procedure for \method{} is formalized in Algorithm~\ref{pseudo:test_eval}. 
Given an initial observation $I_0$ and goal observation $I_g$, the algorithm iteratively generates policy parameters, samples actions, and applies them until either the goal is reached (as determined by the latent distance falling below a threshold $\epsilon$) or a maximum number of steps $T$ is achieved. 
This simple yet effective procedure demonstrates how our parameter-adaptive approach seamlessly integrates into practical robotic control scenarios.

\begin{algorithm}[!h]
  \caption{{\bf \method}: Test‑Time Task Evaluation}
  \begin{tabular}{ll}
    \textbf{Input}:      & Initial observation $I_0$, Goal observation $I_g$ \\
    \textbf{Modules}:    & Encoder $\mathcal{E}$, Policy generation hypernetwork $\mathcal{H}$ \\
    \textbf{Parameters}: & Max steps $T$, Goal completion threshold $\epsilon$ 
  \end{tabular}
  \begin{algorithmic}[1]
    \State $I_t \leftarrow I_0$, $t \leftarrow 0$, $\text{done} \leftarrow \text{false}$
    \While{$t < T$ and not \text{done}}
      \State $\theta \leftarrow \mathcal{H}(\mathcal{E}(I_t), \mathcal{E}(I_g))$      \Comment{Generate policy weights}
      \State $\hat a_t \leftarrow \pi(o_{t-L:t};\theta)$                \Comment{Sample action}
      \State $I_{t+1}, \text{done} \leftarrow \mathrm{Env}(\hat a_t)$     \Comment{Apply action and Env. interaction}
      \If{$d(\mathcal{E}(I_{t+1}),\mathcal{E}(I_g)) < \epsilon$ or done}
        \State \Return SUCCESS
      \EndIf
      \State $I_t \leftarrow I_{t+1}$, $t \leftarrow t + 1$
    \EndWhile
    \State \Return TIMEOUT
  \end{algorithmic}
  \label{pseudo:test_eval}
\end{algorithm}

\section{Experiments}
In this section, we present a comprehensive experimental evaluation across a suite of simulated and real‑robot manipulation tasks designed to address the following questions:
1) How effectively does \method{}'s parameter-adaptive approach generate successful policies for diverse manipulation tasks?
2) To what extent does our latent space shaping enhance the performance of the hypernetwork for goal-conditioned policy generation?
3) How does \method{} compare with conventional goal-conditioned methods and alternative representation learning approaches?
Through extensive empirical analysis, we validate the effectiveness of our parameter-adaptive framework.

\subsection{Experiment Setup} 
\textbf{Simulation Environment.} 
We evaluate our approach using Robosuite, 
a comprehensive robotics benchmark designed 
for both short and long-horizon manipulation tasks~\cite{mandlekar2023mimicgen,zhu2020robosuite}. 
This framework provides a standardized suite of environments, 
from which we select multiple contact-rich tabletop manipulation tasks: 
coffee manipulation, threading, mug cleanup, nut assembly, three-piece assembly, and several long-horizon tasks including coffee preparation and kitchen manipulation. 
To assess robustness across varying initial conditions, we use three difficulty levels ($\text{d}_0$, $\text{d}_1$, $\text{d}_2$), where higher indices correspond to increased environmental variability, 
particularly in object pose initialization (position and orientation). 
Each experimental environment features a robotic manipulator positioned adjacent to a workspace containing task-specific manipulable objects.

\textbf{Training Protocol.} Our approach follows the behavior cloning paradigm, utilizing a dataset based on MimicGen~\cite{mandlekar2023mimicgen}. 
For each task, we employ a training dataset of 950 demonstrations, where each timestep comprises front-view RGB images ($128 \times 128$ resolution), robot proprioceptive states $s_t \in \mathcal{S}$, and corresponding ground-truth actions $a_t \in \mathcal{A}$. The training procedure employs the Adam optimizer~\cite{kingma2014adam} with a cosine learning rate schedule~\cite{loshchilov2016sgdr}. We initialize the learning rate at $5\times10^{-4}$ and maintain uniform loss balancing coefficients ($\lambda_i = 1$ for all components). Our model is trained for 500 epochs with a batch size of 256, by default. Detailed implementation and training protocols are provided in the Sec.~\ref{apx:learning_details}.

\subsection{Main Results}

\textbf{Baselines.} We compare our parameter-adaptive approach against state-of-the-art goal-conditioned methods that use fixed network parameters. 
All methods are trained on pre-collected demonstrations from MimicGen~\cite{mandlekar2023mimicgen} and modified to operate with a single future image as the goal specification for fair comparison:
\begin{itemize}[leftmargin=0.8em]
    \item \textbf{GCBC}~\cite{lynch2020learning, emmons2021rvs}: Goal-Conditioned Behavioral Cloning concatenates current and goal observations as input to a fixed policy network, learning a direct mapping from this concatenated representation to actions through supervised learning on demonstration data.
    \item \textbf{Play-LMP}~\cite{lynch2020learning}: Play-supervised Latent Motor Plans learns a latent plan space from demonstration data, then trains a fixed-parameter policy conditioned on both the current state and the inferred latent plan for the specified goal.
    \item \textbf{C-BeT}~\cite{cui2022play}: Conditional Behavior Transformer uses self-attention to compress observation history into a latent representation, which is combined with the goal state to condition a fixed-parameter transformer that predicts actions.
    \item \textbf{MimicPlay}~\cite{wang2023mimicplay}: MimicPlay is a self-supervised approach that learns general robotic skills from unstructured teleoperation data, which consists of continuous sequences of observations and actions from a human video. For our experiments, this method is adapted into a goal-image-conditioned policy, with implementation details provided in the Appendix.
\end{itemize}

\begin{table*}[t]
\centering
\setlength{\tabcolsep}{0.008\linewidth}
\caption{\textbf{Comparison with state-of-the-art goal-conditioned methods.} 
Success rates (higher is better) are computed over 50 rollouts across various manipulation tasks with increasing difficulty levels (d0-d2). 
The experimental setup uses only two historical observations and a single goal image, representing a practical deployment scenario. 
Our method consistently outperforms conventional fixed-parameter approaches, demonstrating the effectiveness of dynamically generating policy parameters based on goal specifications.
}
\resizebox{\textwidth}{!}{
\begin{tabular}{l|cccc|ccc|cccc|cccc|c|c}
\toprule
\multirow{2}{*}{Method} &
\multicolumn{4}{c|}{Coffee $\uparrow$} & 
\multicolumn{3}{c|}{Mug-cleanup $\uparrow$} &
\multicolumn{4}{c|}{Three piece Assemb. $\uparrow$} &
\multicolumn{4}{c|}{Threading $\uparrow$} &
\multicolumn{1}{c|}{Nut Assemb. $\uparrow$} &
\cellcolor{gray!30} \\
& d0 & d1 & d2 & \cellcolor{gray!30}Avg. & d0 & d1 & \cellcolor{gray!30}Avg. & d0 & d1 & d2 &
\cellcolor{gray!30}Avg. & d0 & d1 & d2 & \cellcolor{gray!30}Avg. & d0 & \cellcolor{gray!30}\multirow{-2}*{Avg.} \\
\midrule
GCBC        & 0.00 & 0.00 & 0.00 & \cellcolor{gray!30}0.00 & 0.00 & 0.00 &  \cellcolor{gray!30}0.00 & 0.00 & 0.00 & 0.00 & \cellcolor{gray!30}0.00& 0.00 & 0.00 & 0.00 & \cellcolor{gray!30}0.00 & 0.00 &\cellcolor{gray!30}0.00 \\
Play-LMP    & 0.00 & 0.00 & 0.00 & \cellcolor{gray!30}0.00 & 0.00 & 0.00 &  \cellcolor{gray!30}0.00 & 0.00 & 0.00 & 0.00 & \cellcolor{gray!30}0.00& 0.00 & 0.00 & 0.00 & \cellcolor{gray!30}0.00 & 0.00 & \cellcolor{gray!30}0.00 \\
MimicPlay & 0.28 & 0.28 & 0.16 & \cellcolor{gray!30} 0.24 & 0.26 & 0.06 &  \cellcolor{gray!30} 0.16 & 0.06 & 0.06 & 0.00 & \cellcolor{gray!30} 0.04 & 0.18 & 0.02 & 0.00 & \cellcolor{gray!30} 0.07 & 0.03 & \cellcolor{gray!30} 0.12 \\
C-BeT   & 0.92 & 0.00 & \textbf{0.74} & \cellcolor{gray!30} 0.55  & 0.30 & \textbf{0.50} &  \cellcolor{gray!30} 0.40 & 0.00 & 0.02 & 0.00 & \cellcolor{gray!30} 0.01 & 0.62 & 0.22 & 0.12 & \cellcolor{gray!30} 0.32 & 0.34 & \cellcolor{gray!30} 0.32 \\

\midrule
Ours     & \textbf{0.94} & \textbf{0.76} & 0.62 & \cellcolor{gray!30}\textbf{0.77} & \textbf{0.78} & 0.46 & \cellcolor{gray!30} \textbf{0.62} & \textbf{0.52} & \textbf{0.20} & \textbf{0.04} & \cellcolor{gray!30}\textbf{0.25} & \textbf{0.82} & \textbf{0.32} & \textbf{0.24} & \cellcolor{gray!30} \textbf{0.46} & \textbf{0.55} & \cellcolor{gray!30}\textbf{0.52} \\
\bottomrule
\end{tabular}%
}
\label{tab:main_result}
\end{table*}

\begin{table}[t]
  \centering
  \setlength{\tabcolsep}{0.01\linewidth}
  \begin{minipage}[t]{0.48\textwidth}
  \captionof{table}{\textbf{Long-horizon task performance.} Our parameter-adaptive approach excels in complex sequential tasks, outperforming fixed-parameter methods across difficulty levels.}
    \vspace{0.5ex}
    \resizebox{\textwidth}{!}{
    {\small   
      \begin{tabular}{l|ccc|ccc|c}
        \toprule
        \multirow{2}{*}{Method} & \multicolumn{3}{c|}{Coffee Preparation$\uparrow$} & \multicolumn{3}{c|}{Kitchen$\uparrow$} & \cellcolor{gray!30} \\
        & d0 & d1 & \cellcolor{gray!30}Avg. & d0 & d1 & \cellcolor{gray!30}Avg. & \cellcolor{gray!30}\multirow{-2}*{Avg.}\\
        \midrule
        GCBC & 0.00 & 0.00 & \cellcolor{gray!30}0.00 & 0.00 & 0.00 & \cellcolor{gray!30}0.00 &  \cellcolor{gray!30}0.00 \\
        Play-LMP & 0.00 & 0.00 & \cellcolor{gray!30}0.00 & 0.00 & 0.00 & \cellcolor{gray!30}0.00  & \cellcolor{gray!30}0.00 \\
        MimicPlay & 0.34 & 0.00 & \cellcolor{gray!30}0.17 & 0.86 & 0.18 & \cellcolor{gray!30}0.52  & \cellcolor{gray!30}0.35 \\
        C-BeT & \textbf{0.82} & 0.04 & \cellcolor{gray!30}0.43 & 0.78 & 0.70 & \cellcolor{gray!30}0.74 &  \cellcolor{gray!30}0.59 \\
        \midrule
        Ours & 0.80 & \textbf{0.50} & \cellcolor{gray!30}\textbf{0.65} & \textbf{1.00} & \textbf{0.80} & \cellcolor{gray!30}\textbf{0.90} & \cellcolor{gray!30}\textbf{0.78} \\
        \bottomrule
        \end{tabular}
    }
    }
    \label{tab:long_horizon}
  \end{minipage}%
  \hfill
  \setlength{\tabcolsep}{0.03\linewidth}
  \begin{minipage}[t]{0.48\textwidth}
  \captionof{table}{\textbf{Component ablation analysis.} Latent space shaping significantly enhances performance, while proper distance metrics and training are crucial for robustness.}
    \vspace{0.5ex}
    \resizebox{\textwidth}{!}{
    {\small   
      \begin{tabular}{l|cccc}
        \toprule
        \multirow{2}{*}{Method} &
        \multicolumn{4}{c}{Coffee $\uparrow$} \\
         & d0 & d1 & d2 & \cellcolor{gray!30}Avg. \\
        \midrule
        Ours(uf. at epoch 0) & 0.92 & 0.00 & 0.62 & \cellcolor{gray!30} 0.51 \\
        Ours(w/o shaping) & 0.92 & 0.00 & 0.00 & \cellcolor{gray!30} 0.31 \\
        Ours(dist$\leftrightarrow$start img.) & 0.50 & 0.52 & 0.32 & \cellcolor{gray!30} 0.45 \\
        Ours(cos dist) & \textbf{0.94} & 0.36 & 0.48 & 0.59\cellcolor{gray!30} \\
        C-BeT(w/ shaping) & 0.80 & 0.64 & \textbf{0.64} & 0.69\cellcolor{gray!30} \\
        \midrule
        Ours & \textbf{0.94} & \textbf{0.76} & 0.62  & \cellcolor{gray!30}\textbf{0.77} \\
        \bottomrule
        \end{tabular}
    }
    }
    \label{tab:ablation}
  \end{minipage}
\end{table}

\noindent \textbf{Evaluation Metrics.}
We evaluate performance using task completion success rates over 50 independent rollouts with \textbf{randomly initialized}, previously unseen environmental configurations. We impose maximum trajectory lengths of  $T=600$ or $800$  steps for contact-rich tasks and $T=1600$ for long-horizon tasks. While our parameter-adaptive approach enables autonomous task completion detection through latent space metrics (Algorithm~\ref{pseudo:test_eval}), we use environment-provided terminal signals for standardized evaluation across all methods. Success is indicated as $\mathcal{S}_i = 1$ if rollout $i$ completes within $T$ steps, and $\mathcal{S}_i = 0$ otherwise.

\noindent \textbf{Quantitative Results.}
Tables~\ref{tab:main_result} and~\ref{tab:long_horizon} present success rates across multi-step and long-horizon tasks, respectively. For each task, success is determined by task-specific criteria provided by the environment -- such as correct object placement, successful insertion, proper assembly configuration, or completion of a sequence of subtasks for long-horizon scenarios. 
Our parameter-adaptive approach outperforms fixed-parameter methods across these diverse evaluation criteria and difficulty levels. This superior performance stems from our hypernetwork's ability to dynamically generate task-specific policy parameters tailored to each goal, resulting in more effective goal-directed behavior. Particularly in high-variability environments (difficulty levels d1-d2), our method demonstrates greater robustness and adaptability compared to conventional approaches -- highlighting the advantage of having policy parameters explicitly conditioned on goals rather than using fixed parameters for all scenarios.

\noindent \textbf{Analysis of Likelihood-Based Baselines.}
We diagnose the poor performance of GCBC and Play-LMP as a fundamental issue of their learning objective, not implementation. These methods, adapted from reputable third-party code, aim to maximize the log-likelihood of expert actions. We found this leads to severe overfitting on the training data, evidenced by a large gap between low training loss and high validation loss. Such memorization-based learning fails to generalize to the subtle variations present in our high-precision test scenarios. This limitation of likelihood-based models in complex settings is corroborated by prior work~\cite{cui2022play}. In contrast, our hypernetwork's design imposes a beneficial structural bias: by emulating an optimization process to generate parameters, it is incentivized to learn a functional mapping from goal to policy, ensuring better generalization and avoiding the overfitting issues that plague the baselines.

\begin{wraptable}{r}{7cm}
\vspace{-11pt}
\caption{
\textbf{Ablating hypernetwork architectures.} Our method with standard initialization is compared against HyperZero variants stabilized with enhanced initializations.}
\vspace{-8pt}
\label{tab:hyperzero}
\begin{center}
\begin{small}
\begin{tabular}{l|cccc}
\toprule
\multirow{2}{*}{Method} & \multicolumn{4}{c}{Coffee Task (\%) $\uparrow$} \\
& d0 & d1 & d2 & \cellcolor{gray!30}Avg. \\
\midrule
HyperZero + \texttt{ScalarInit}    & 16 & 18 & 14 & \cellcolor{gray!30}16 \\
HyperZero + \texttt{Bias-Init} & 30 & 18 & 0  & \cellcolor{gray!30}16 \\
\midrule
\textbf{Ours (Standard Init.)} & \textbf{94} & \textbf{76} & \textbf{62} & \cellcolor{gray!30}\textbf{77} \\
\bottomrule
\end{tabular}
\end{small}
\end{center}
\vspace{-10pt}
\end{wraptable}

\subsection{Ablation Study} 
\textbf{Hypernetwork Architecture Analysis.}
We evaluate our optimization-inspired hypernetwork design against HyperZero~\cite{rezaei2023hypernetworks}, a prominent alternative architecture. Architecturally, HyperZero encodes conditioning information into a meta-embedding that is then transformed to produce the parameters for the target policy network. Because this direct mapping can produce parameters with a numerical range misaligned with that of an optimally trained network, it often requires special initialization to stabilize training. 
To ensure a stable and robust comparison, we therefore implemented two enhanced initialization schemes for HyperZero. The first, \texttt{ScalarInit}, introduces a learnable scalar to control the initial scale of the hypernetwork's output. The second, \texttt{Bias-Init}~\cite{beck2023hypernetworks}, is designed for high-dimensional conditioning inputs and constrains the parameter range by incorporating learnable biases alongside weights that are initialized to zero. 
As shown in Table~\ref{tab:hyperzero}, even with these stabilization techniques, our architecture, which requires no special initialization, vastly outperforms HyperZero. This demonstrates that the iterative refinement mechanism in our design is inherently more effective at capturing the relationship between goals and appropriate policy parameters. These results highlight the critical role of architectural choice in developing robust parameter-adaptive policies. More details can be found in Sec.~\ref{apx:hpzero}.

\begin{wrapfigure}{r}{0.5\textwidth}
  \centering
  \vspace{-9pt}
  \includegraphics[width=0.95\linewidth]{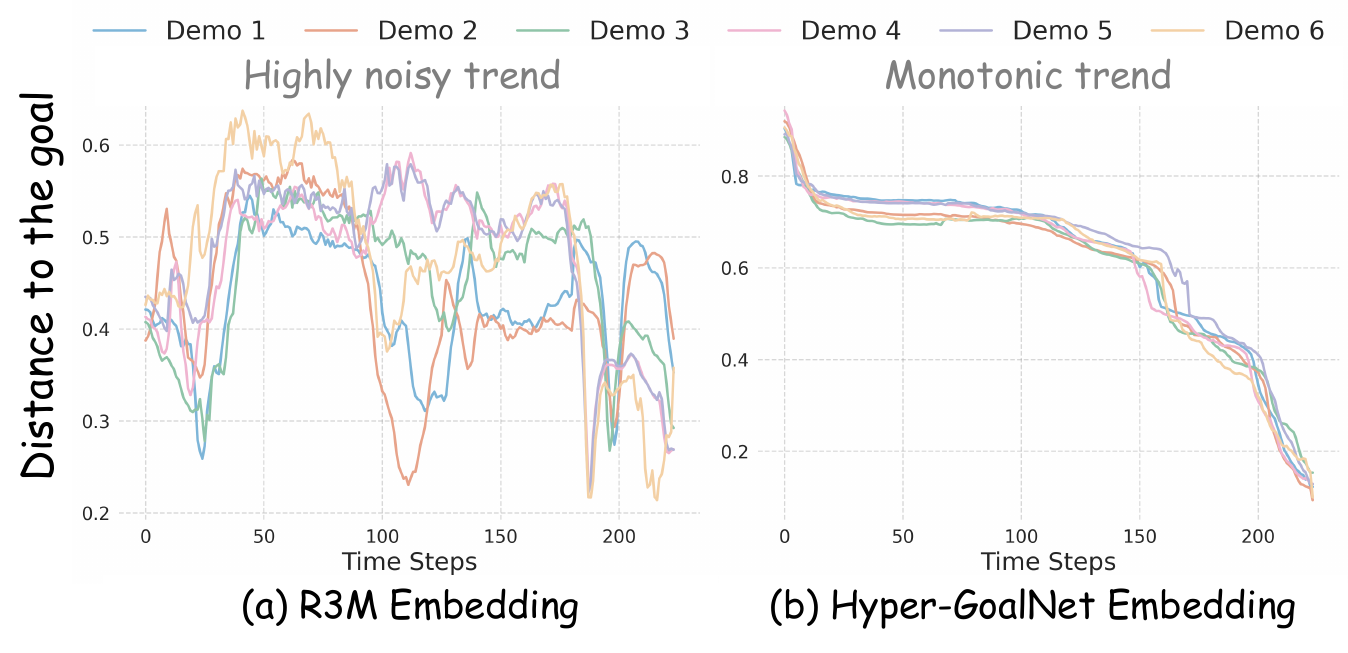}
  \caption{
   Comparison of (a) unshaped R3M embeddings versus (b) our shaped latent space, showing $L_2$ distances to goal states along multiple trajectories. 
   Our shaping creates consistent monotonic decreases in distance-to-goal, facilitating more effective parameter generation.}
  \label{fig:effect_shaping}
  \vspace{-12pt}
\end{wrapfigure}

\textbf{Latent Space Shaping Analysis.}
Table~\ref{tab:ablation} shows that 
proper latent space shaping is critical for our parameter-adaptive framework. 
Removing shaping (``Ours w/o shaping'') severely degrades performance on harder difficulty levels, 
while our specific choices of using goal-relative distances and Euclidean metrics prove superior to alternatives. 
Figure~\ref{fig:effect_shaping} visually confirms how our approach creates more consistent monotonic progression toward goals compared to unshaped representations.
Notably, applying our shaping techniques 
to the baseline C-BeT improves its performance, 
yet its performance lags compared to ours, 
which in turn signifies the importance 
of the parameter-adaptive framework 
as well as the synergy between policy generation 
and latent shaping. 
Our training strategy also matters -- unfreezing the R3M visual encoder~\cite{nair2022r3m} only after 20 epochs ensures stable parameter generation. 
These results validate our insight that latent spaces should reflect physical progression toward goals to effectively support parameter-adaptive policy learning.

\textbf{Visualization.} 
Figures~\ref{fig:effect_shaping} and~\ref{fig:exp_goalreaching} illustrate how our latent space shaping creates representations that directly benefit our parameter-adaptive approach. The plot shows the latent distance to the goal (y-axis) over the execution timesteps of a robotic task rollout (x-axis), where later steps are progressively closer to the goal. Unlike R3M embeddings which show significant fluctuations, our method produces consistently monotonic distance reductions toward goal states. 
This structured latent space offers two key advantages for our hypernetwork: (1) it provides clearer signals for generating appropriate policy parameters as the agent progresses toward goals, and (2) it enables reliable autonomous detection of task completion based on latent distances. The visualization confirms that our combined predictive modeling and distance constraints create optimal representations for goal-conditioned parameter generation, enhancing both performance and interpretability.

\begin{wraptable}{r}{7.6cm}
\vspace{-10pt}
\caption{
\textbf{Quantitative validation of autonomous goal completion detection.} Our method's latent distance-based success rate (Auto SR) is compared against the environment's ground truth (Env SR) on the Coffee tasks. High Accuracy and Recall validate its reliability.}
\vspace{-8pt}
\label{tab:goal_detection}
\begin{center}
\begin{small}
\begin{tabular}{lcccc}
\toprule
Task & Auto SR & Env SR & Accuracy & Recall \\
\midrule
D0 & 0.96 & 0.94 & 94\% & 98\% \\
D1 & 0.78 & 0.76 & 90\% & 95\% \\
D2 & 0.74 & 0.62 & 76\% & 90\% \\
\midrule
\textbf{Mean} & -- & -- & \textbf{86.6\%} & \textbf{94.3\%} \\
\bottomrule
\end{tabular}
\end{small}
\end{center}
\vspace{-10pt}
\end{wraptable}

\textbf{Goal Completion Detection.}
To substantiate our claim that the shaped latent space enables autonomous goal completion detection, we supplement the qualitative evidence from Figures~\ref{fig:effect_shaping}, \ref{fig:more_vis_1}, and \ref{fig:more_vis_2}. We evaluate this capability by comparing our autonomous detection, where success is determined by a latent distance threshold (Auto SR), against the environment's ground-truth signal (Env SR). Table~\ref{tab:goal_detection} presents the results using two key metrics: \textit{Accuracy}, which measures the agreement between the two signals, and \textit{Recall}, which measures our method's ability to identify true successes reported by the environment. The strong alignment, evidenced by an average accuracy of 86.6\% and recall of 94.3\%, provides strong empirical evidence that our latent-distance-based approach is a reliable autonomous completion detector.

\begin{wraptable}{r}{7cm}
\vspace{-5pt}
\caption{
\textbf{Real‑robot experiment results.} (successes/total trials).}
\vspace{-8pt}
\label{tab:real_exp}
\begin{center}
\begin{small}
\begin{tabular}{cccccc}
    \toprule
    Method & Pickplace & Pull & Stack & Sweep \\
    \midrule
    GCBC & 0/15 & 0/15 & 0/15 & 2/15 \\
    Play-LMP & 0/15 & 0/15 & 0/15 & 5/15 \\
    C-BeT  & 2/15 & 6/15 & 5/15 & 8/15 \\
    \midrule
    Ours  & 14/15 & 15/15 & 14/15 & 15/15 \\
    \bottomrule
\end{tabular}
\end{small}
\end{center}
\vspace{-20pt}
\end{wraptable}

\subsection{Real Robot Experiments}
We validate our parameter-adaptive approach on physical hardware using the Realman Robotics Platform, featuring a 7-DoF manipulator with a 1-DoF parallel gripper (Figure~\ref{fig:workspace}). We evaluate four diverse manipulation tasks -- sweep, pick\&place, pull, and stack -- with 15 trials per task. Due to hardware constraints limiting control to joint angles without end-effector pose information, we exclude MimicPlay, which requires precise 3D end-effector trajectories.

As shown in Table~\ref{tab:real_exp}, conventional fixed-parameter approaches struggle significantly in real-world conditions where environmental noise, perception uncertainties, and imperfect demonstrations create substantial challenges. In contrast, our method maintains high success rates across all tasks, including those requiring precise contact-rich interactions. This real-world performance gap highlights a key advantage of our parameter-adaptive approach: by dynamically generating task-specific policy parameters based on goal images, our method better adapts to real-world variations and demonstration imperfections that weren't encountered during training. Detailed experimental protocols are provided in the Sec.~\ref{apx:real_setting}.
\section{Discussion}
\textbf{Conclusion.}
Our parameter-adaptive approach represents an effective move in goal-conditioned policy learning by dynamically generating policy parameters based on goal information rather than using fixed parameters with conditioning. 
The consistent performance improvements across tasks demonstrate that ``how'' observations should be processed is inherently dependent on the goal specification. 
Our latent space shaping techniques prove critical for this architecture -- imposing physical structure and predictive capacity provides clearer signals for the hypernetwork to generate effective policy parameters.
Overall, our results suggest that explicitly modeling the relationship between goals and processing mechanisms offers a promising direction for more flexible and robust robotic control.

\textbf{Limitations.}
Our method's primary limitation is its reliance on a well-structured latent space, which is challenging to form for highly complex tasks and requires demonstration data with clear goal progression. This data dependency creates a key failure mode: out-of-distribution goals can cause the hypernetwork to generate erratic policies. Furthermore, as an offline-trained method, its dynamic generation of parameters for novel goals lacks explicit safety guarantees against unforeseen states, making the integration of robust safety constraints a critical direction for future research.

\begin{ack}
This work is supported by 
the Early Career Scheme 
of the Research Grants Council (RGC) grant \# 27207224,
the HKU-100 Award, 
a donation from the Musketeers Foundation, 
and in part by the 
JC STEM Lab of 
Autonomous Intelligent Systems 
funded by 
The Hong Kong Jockey Club Charities Trust.
\end{ack}

\bibliographystyle{plain}
\bibliography{example}

\clearpage
\section*{Appendix}
In this section, we present supplementary materials detailing the methodological framework and experimental procedures used in this study.

\begin{appendix}

\section{Manipulation Task Details}

\paragraph{Task Details.} Our experimental evaluation was conducted within the Robosuite~\cite{zhu2020robosuite} simulation environment, utilizing the benchmark dataset from Mimicgen~\cite{mandlekar2023mimicgen}. We primarily investigated complex tabletop manipulation tasks that encompass diverse robotic skills. The selected tasks are characterized as follows:

\begin{itemize}
\item \textbf{Coffee:} A multi-step manipulation task requiring precise object handling, where the robot must grasp a coffee capsule, insert it into the designated slot of the coffee machine, and securely close the machine's lid.
\item \textbf{Mug cleanup:} A sequential task involving both articulated object interaction and object placement. The robot must coordinate drawer manipulation and object transportation, culminating in storage of a mug.
\item \textbf{Three piece assembly:} A multi-step assembly task demanding spatial reasoning and precise manipulation. The robot must stack three components in a specific sequence to achieve compact assembly configuration.
\item \textbf{Threading:} A high-precision manipulation task requiring fine motor control. The robot must accurately orient and manipulate a needle for successful insertion through a minimal aperture.
\item \textbf{Nut Assembly:} A manipulation task involving precise grip control and spatial alignment for successful mechanical assembly. In this task, the success rates for the two nuts are measured separately, and the overall success rate is then calculated.
\item \textbf{Coffee Preparation:} An extended sequential task combining multiple sub-goals, including cup positioning, drawer manipulation, capsule retrieval, and coffee maker operation, culminating in a fully prepared coffee setup.
\item \textbf{Kitchen:} A complex sequence involving appliance interaction, object manipulation, and spatial reasoning. The task includes stove operation, cookware handling, and precise object placement.
\end{itemize}

\begin{figure*}[t]
  \centering
  \includegraphics[width=\linewidth]{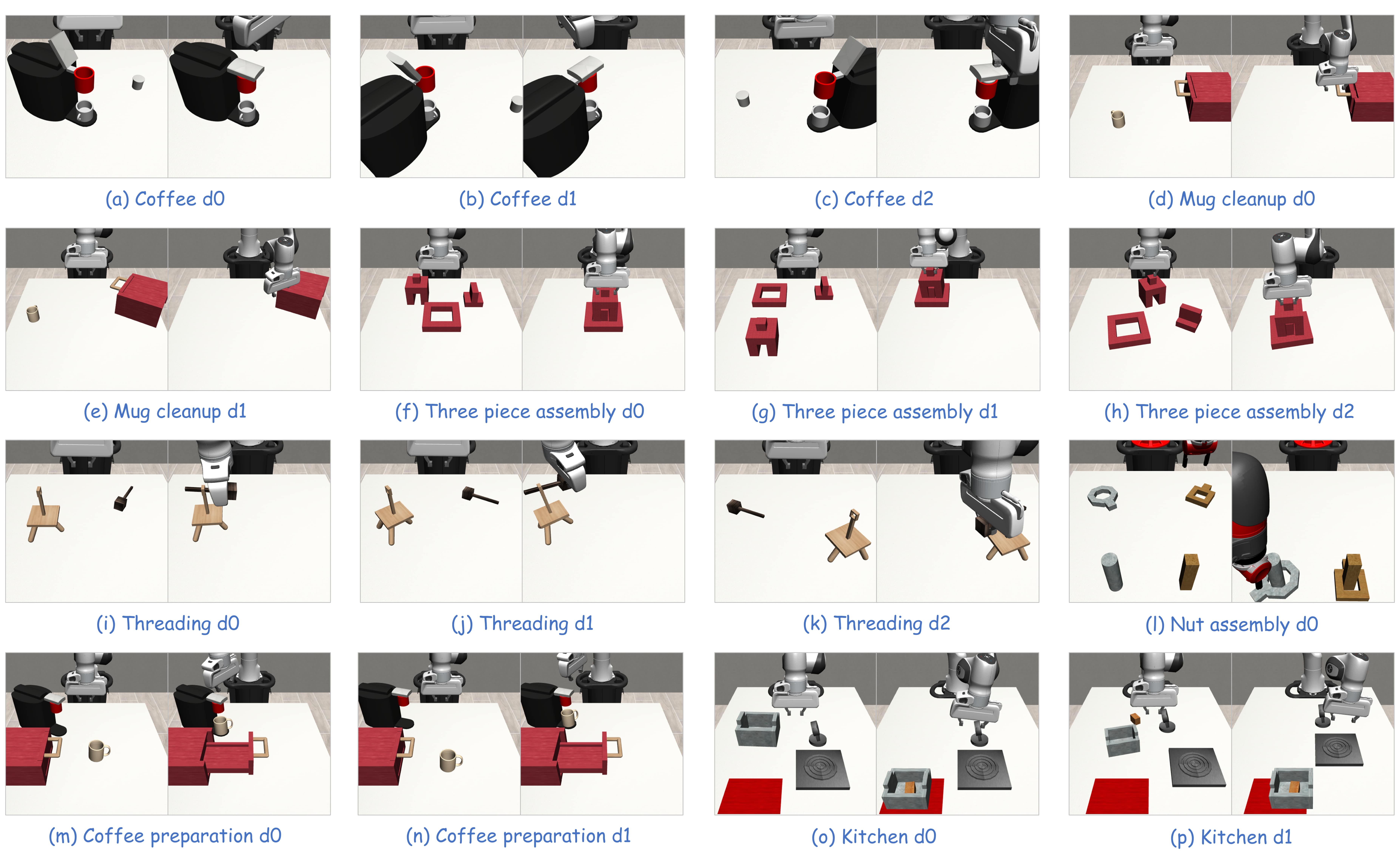}
   \caption{Key phases of diverse manipulation tasks in experimental evaluation.}
   \label{fig:task_desp}
\end{figure*}

These tasks are specifically selected for their comprehensive representation of challenging robotic manipulation scenarios, featuring contact-rich interactions, precise object manipulation, and complex multi-step sequences. Each task requires a combination of skills including spatial reasoning, and sequential decision-making. The visual representation of these manipulation tasks and their key phases are illustrated in Figure~\ref{fig:task_desp}.

\paragraph{Data Processing and Observation Space.} The demonstration data from Mimicgen is initially preprocessed by segmenting complete demonstrations into trajectory subsequences to facilitate learning. Our framework implements a constrained observation context with a length of 2, including front-view RGB images and the agent's proprioceptive state information. This design choice means that the agent's decision-making process is based solely on the current frame and one historical frame, deliberately limiting the temporal horizon to enhance real-world applicability. Given our focus on goal-conditioned policy learning, the observation space of the hypernetwork is augmented with a single RGB goal image representing a feasible target state. The input modalities are structured as follows:

\begin{itemize}
\item Visual observations: RGB images with dimensions 128×128 pixels for both contextual and goal representations.
\item Proprioceptive state: A compact 9-dimensional vector encoding essential agent state information.
\end{itemize}

This deliberately constrained observation space creates a partially observable environment that closely aligns with real-world robotics scenarios, where complete state information is rarely available. While this design choice enhances the practical applicability of our approach, it also introduces significant challenges:

\begin{itemize}
\item Single-goal scenarios necessitate hypernetwork architectures to efficiently handle the demanding requirements of goal-conditioned policy generation.
\item Limited temporal context requiring efficient use of historical information.
\item Partial observability demanding robust state estimation and feature extraction.
\item Complex vision-based reasoning with constrained visual information.
\end{itemize}

Such challenging conditions serve to validate our method's effectiveness under realistic constraints, demonstrating its potential for real-world deployment.

\paragraph{Goal Specification Strategy.} 

Our approach implements a systematic strategy for goal specification across both training and evaluation phases. During training, we employ a dynamic goal sampling mechanism where the goal image is stochastically selected from future frames within the same demonstration, subsequent to the current timestep. This design offers two key advantages:

\begin{itemize}
\item \textbf{Goal Feasibility:} By sampling from actual demonstration frames, we inherently guarantee the physical feasibility and reachability of the specified goals.
\item \textbf{Goal Diversity:} The random sampling mechanism ensures sufficient variation in goal states, promoting the learning of a robust and generalizable policy.
\end{itemize}

During the evaluation phase, 
the goal specification mechanism leverages the Mimicgen framework to generate feasible goal states, facilitating potential transfer from simulation to physical systems.

\begin{table*}[h]
\centering
\setlength{\tabcolsep}{0.008\linewidth}
\caption{\textbf{Additional quantitative experiments with more baseline methods across different tasks.} $\dagger$ indicates methods using a \textbf{sequence of visual frames} as goals rather than a single goal image, with an \textbf{extended observation length} of 10 frames versus the standard 2. $\ddagger$ indicates methods with access to \textbf{extra} wrist-mounted camera images. The wrist views, extended observation sequence, and extended goal sequences provide richer observational and task-guidance information for augmented baselines. \method{} leverages only one goal image and two front-view observations, indicating a more easy-to-be-applied setup while being more effective.}
\resizebox{\textwidth}{!}{
\begin{tabular}{l|cccc|ccc|cccc|cccc|c|c}
\toprule
\multirow{2}{*}{Method} &
\multicolumn{4}{c|}{Coffee $\uparrow$} & 
\multicolumn{3}{c|}{Mug-cleanup $\uparrow$} &
\multicolumn{4}{c|}{Three piece Assemb. $\uparrow$} &
\multicolumn{4}{c|}{Threading $\uparrow$} &
\multicolumn{1}{c|}{Nut Assemb. $\uparrow$} &
\cellcolor{gray!30} \\
& d0 & d1 & d2 & \cellcolor{gray!30}Avg. & d0 & d1 & \cellcolor{gray!30}Avg. & d0 & d1 & d2 &
\cellcolor{gray!30}Avg. & d0 & d1 & d2 & \cellcolor{gray!30}Avg. & d0 & \cellcolor{gray!30}\multirow{-2}*{Avg.} \\
\midrule
MimicPlay-O$^{\dagger\ddagger}$   & 0.80 & 0.84 & 0.88 & \cellcolor{gray!30} 0.84 & 0.68 & 0.58 &  \cellcolor{gray!30} 0.63 & 0.50 & 0.38 & 0.02 & \cellcolor{gray!30} 0.30 & 0.32 & 0.06 & 0.06 & \cellcolor{gray!30} 0.15 & 0.10 & \cellcolor{gray!30} 0.44 \\
MimicPlay-M$^{\dagger}$   & 0.28 & 0.28 & 0.16 & \cellcolor{gray!30} 0.24 & 0.26 & 0.06 &  \cellcolor{gray!30} 0.16 & 0.06 & 0.06 & 0.00 & \cellcolor{gray!30} 0.04 & 0.18 & 0.02 & 0.00 & \cellcolor{gray!30} 0.07 & 0.03 & \cellcolor{gray!30} 0.12 \\
\midrule
C-BeT   & 0.92 & 0.00 & 0.74 & \cellcolor{gray!30} 0.55  & 0.30 & 0.50 &  \cellcolor{gray!30} 0.40 & 0.00 & 0.02 & 0.00 & \cellcolor{gray!30} 0.01 & 0.62 & 0.22 & 0.12 & \cellcolor{gray!30} 0.32 & 0.34 & \cellcolor{gray!30} 0.32 \\
\midrule

\method{}(G)     & \textbf{0.94} & \textbf{0.60} & \textbf{0.62} & \cellcolor{gray!30}\textbf{0.72} & \textbf{0.72} & \textbf{0.54} & \cellcolor{gray!30} \textbf{0.63} & \textbf{0.46} & \textbf{0.22} & \textbf{0.02} & \cellcolor{gray!30}\textbf{0.23} & \textbf{0.78} & \textbf{0.20} & \textbf{0.18} & \cellcolor{gray!30} \textbf{0.39} & \textbf{0.67} & \cellcolor{gray!30}\textbf{0.50} \\

\method{}    & \textbf{0.94} & \textbf{0.76} & \textbf{0.62} & \cellcolor{gray!30}\textbf{0.77} & \textbf{0.78} & \textbf{0.46} & \cellcolor{gray!30} \textbf{0.62} & \textbf{0.52} & \textbf{0.20} & \textbf{0.04} & \cellcolor{gray!30}\textbf{0.25} & \textbf{0.82} & \textbf{0.32} & \textbf{0.24} & \cellcolor{gray!30} \textbf{0.46} & \textbf{0.55} & \cellcolor{gray!30}\textbf{0.52}\\
\bottomrule
\end{tabular}%
}
\label{tab:add_result}
\end{table*}

\section{Comparison with Baselines}

Given the challenging nature of our experimental setting as shown in Table~\ref{tab:main_result}, existing baseline methods demonstrate limited success in task completion. To ensure comprehensive evaluation, we introduce modified versions of existing approaches and additional baseline methods adapted for our scenario. The following section details these enhanced baseline implementations and their comparative performance.

We also conducted additional comparative experiments with BeT~\cite{shafiullah2022behavior}, C-BeT~\cite{cui2022play}, and MimicPlay~\cite{wang2023mimicplay} (implemented in both its original setting and a modified setting). And a variant of our method, \method{}(G) is also introduced. Experiments were performed across all 16 MimicGen tasks, with results for contact-rich tasks and long-horizon tasks presented in Tables~\ref{tab:add_result} and \ref{tab:add_long_horizon}, respectively. 
Please note that the success rates reported in Table~\ref{tab:add_long_horizon} reflect a modified evaluation criterion in the simulation environment, resulting in slight variations from the kitchen task results presented in Table~\ref{tab:main_result}.
The implementation details are explained below.

\paragraph{Details about the Baselines.} We selected four task-specific baselines and reimplemented them under settings generally consistent with \method{}. The reimplementation details are as follows.

\begin{itemize}
    \item \textbf{GCBC}~\cite{lynch2020learning, emmons2021rvs}: Goal-Conditioned Behavioral Cloning (GCBC) is the most general framework for learning goal-conditioned policies. It consists of a perception module, a visual encoder, and a RNN-based goal-conditioned policy module. GCBC takes in a 9-dimensional proprioceptive state, a current front-view RGB image, and a goal RGB image as input and predicts the action distribution to transfer the current state to the goal state. The model is trained end-to-end with the objective of maximizing the log-likelihood of the ground-truth action in the predicted distribution. The observation sequence length and the predicted action sequence length are both restricted to 5 steps.\\
    \item \textbf{Play-LMP}~\cite{lynch2020learning}: Play-Supervised Latent Motor Plans (Play-LMP) builds upon the foundation of GCBC, aiming to learn reusable plan representations and task-agnostic control from play data. Play-LMP consists of three main components: 1) Plan recognition module: maps the input sequence to a distribution in the latent plan space. 2) Plan proposal module: generates multiple conditional prior solutions based on the current and goal states. 3) Plan and goal-conditioned policy: predicts actions conditioned on the current state, goal state, and a latent plan sampled from the plan proposals. Similar to GCBC, both the observation sequence length and the predicted action sequence length are restricted to 5 steps. The model is trained end-to-end.\\
    \item \textbf{MimicPlay}~\cite{wang2023mimicplay}: MimicPlay employs a hierarchical learning framework consisting of two training stages. In the high-level training stage, the model takes the robot's end-effector pose, along with the current visual observation and goal observation, as input to predict the future pose trajectory of the robot's end-effector. This component is referred to as the high-level planner. In the low-level training stage, the high-level planner with the best validation performance from the previous stage is loaded and its parameters are frozen. The model then continues training using a 9-dimensional robot proprioceptive state and visual observations (both current and goal) as input to predict the robot's actions.
    
    Please note that in the original MimicPlay low-level training setup, in addition to the current front-view RGB image, a wrist-mounted RGB image is also used as input, which may contribute to its higher success rate. To ensure a fair comparison with our method, we modify this setup by \textit{replacing} the wrist-mounted image with a duplicate front-view image during the low-level training process. Additionally, we adopt the same goal-specified strategy as described above, rather than providing the entire prompt video as used in MimicPlay's original test-time evaluation setting.
\end{itemize}

\paragraph{\method{}(G)} We introduce another variant of our method, \method{}(G), where the hypernetwork backbone takes only a single goal image as input, without requiring the current image during both training and testing phases. All other settings remain unchanged. During inference, \method{}(G) generates the weights for the lightweight target policy only once and maintains them fixed during rollouts, resulting in improved computational efficiency. As shown in Table~\ref{tab:add_result} and Table~\ref{tab:add_long_horizon}, \method{}(G) still outperforms the baselines while utilizing a much smaller lightweight policy network, demonstrating both the effectiveness and efficiency of our approach.

\paragraph{Comparison with BeT~\cite{shafiullah2022behavior}.} We conducted comparative experiments with Behavior Transformer (BeT), a state-of-the-art approach for multi-modal behavioral learning. BeT employs k-means clustering to discretize continuous actions and utilizes transformers to model categorical distributions across action bins, incorporating an action correction head to refine discretized actions into continuous ones. Despite not being explicitly designed as a goal-conditioned policy, BeT has emerged as a robust baseline in current robot learning literature. Given that the original BeT implementation for the Franka Kitchen task was limited to state space observations and lacked compatibility with Robosuite tasks, we enhanced its architecture by incorporating a pretrained image encoder~\cite{nair2022r3m}. To strengthen the baseline comparison, we augmented BeT with additional wrist camera observations—a feature absent in our method—and extended the context length from 2 (used in our approach) to 4, which typically facilitates more effective policy learning for BeT. Consequently, BeT acquires more image observations per timestep than our method, while maintaining the same image resolution of 128×128 pixels.

\begin{table}[t]
\centering
\setlength{\tabcolsep}{0.011\linewidth}
\caption{\textbf{Additional evaluation on long-horizon tasks.} 
$\dagger$ indicates methods using a \textbf{sequence of visual frames} as goals rather than a single goal image, with an \textbf{extended observation length} of 10 frames versus the standard 2. $\ddagger$ indicates methods with access to \textbf{extra} wrist-mounted camera images. The wrist views, extended observation sequence, and extended goal sequences provide richer observational and task-guidance information for augmented baselines.
Ours achieves similar performance compared to the baseline with access to wrist view images, extended observation sequences and a sequence of goal images, which demonstrates the effectiveness of our method in achieving long-horizon tasks with much less guidance information (effort).}
\begin{tabular}{l|ccc|ccc|c}
\toprule
\multirow{2}{*}{Method} & \multicolumn{3}{c|}{Coffee Prep.$\uparrow$} & \multicolumn{3}{c|}{Kitchen$\uparrow$} & \cellcolor{gray!30} \\
& d0 & d1 & \cellcolor{gray!30}Avg. & d0 & d1 & \cellcolor{gray!30}Avg. & \cellcolor{gray!30}\multirow{-2}*{Avg.}\\
\midrule
MimicPlay-O$^{\dagger\ddagger}$ & 0.86 & 0.68 & \cellcolor{gray!30}0.77 & 1.00 & 0.70 & \cellcolor{gray!30}0.85 &  \cellcolor{gray!30}0.81 \\
MimicPlay-M$^\dagger$ & 0.34 & 0.00 & \cellcolor{gray!30}0.17 & 0.86 & 0.18 & \cellcolor{gray!30}0.52  & \cellcolor{gray!30}0.35 \\
\midrule
C-BeT & 0.82 & 0.04 & \cellcolor{gray!30}0.43 & 0.78 & 0.70 & \cellcolor{gray!30}0.74 &  \cellcolor{gray!30}0.59 \\
\midrule
\method{}(G) & \textbf{0.80} & \textbf{0.50} & \cellcolor{gray!30}\textbf{0.65} & \textbf{1.00} & \textbf{0.88} & \cellcolor{gray!30}\textbf{0.94} & \cellcolor{gray!30}\textbf{0.80} \\

\method{} & \textbf{0.80} & \textbf{0.50} & \cellcolor{gray!30}\textbf{0.65} & \textbf{1.00} & \textbf{0.80} & \cellcolor{gray!30}\textbf{0.90} & \cellcolor{gray!30}\textbf{0.78} \\
\bottomrule
\end{tabular}
\label{tab:add_long_horizon}
\end{table}

\begin{table}[t]
\centering
\setlength{\tabcolsep}{0.03\linewidth}
\caption{\textbf{Comparison with augmented BeT.} $\dagger$ represents \textbf{extended} context length and \textbf{additional} wrist-view images. Although BeT is not originally designed as a goal-conditioned policy method, its augmented version serves as a strong baseline. Our method still outperforms the augmented BeT.
}
\begin{tabular}{l|c|c|c|c}
\toprule
Method & Cof. d0 $\uparrow$ & Cof. d2 $\uparrow$ & Mug. d1 $\uparrow$ & \cellcolor{gray!30}Avg. \\
\midrule
BeT$^\dagger$ & 0.66 & 0.42 & 0.26 & \cellcolor{gray!30}0.45 \\
Ours & \textbf{0.94} & \textbf{0.62} & \textbf{0.46} & \cellcolor{gray!30}\textbf{0.67} \\
\bottomrule
\end{tabular}
\label{tab:bet}
\end{table}

Table~\ref{tab:bet} presents experimental results for some randomly selected tasks, where Cof. d0 and Cof. d2 represent Coffee d0 and Coffee d2 tasks, respectively, and Mug. d1 denotes the Mug cleanup d1 task. Notably, despite utilizing reduced contextual information, our method demonstrates superior performance across all tasks in terms of success rates. These results substantiate our method's robust capability in behavior learning, even under more constrained observational conditions.

\paragraph{Comparison with C-BeT~\cite{cui2022play}.} Conditional Behavior Transformer (C-BeT) is a goal-conditioned version of BeT, a behavior prediction model that compresses an agent's observation history and the goal state into a compact latent representation using self-attention, which is then transformed along with discrete action representations to efficiently predict the agent's future behaviors. 
To ensure fair comparison, we configured C-BeT with identical observation settings to our method, maintaining a context length of 2 and utilizing a single goal image. The comparative results are presented in Table~\ref{tab:add_result} and Table~\ref{tab:add_long_horizon}.

\paragraph{Comparison with Original MimicPlay (MimicPlay-O).}In the original MimicPlay experiment setting, we \textit{retained} the wrist image as an input during the low-level training stage. For the test-time evaluation process, we provided the pretrained model with prompt videos in HDF5 format generated by MimicGen. In contrast to our approach, MimicPlay-O has access to wrist-mounted camera images, uses \textbf{an extended observation length of 10 frames} instead of 2, and utilizes \textbf{a sequence of visual frames as goals}, rather than a single goal image configuration.

\paragraph{Comparison with another Modified MimicPlay (MimicPlay-M).} In this experimental setting, we \textit{removed} the wrist image as an input during the low-level training stage, since the wrist image is not easy to obtain for goal specification. For the test-time evaluation process, we provided the pretrained model with prompt videos in HDF5 format generated by MimicGen. Please note that this experimental setup slightly differs from the one in our baseline setting. In contrast to our approach, 
MimicPlay-M uses an extended observation length of 10 frames instead of 2, and utilizes a sequence of visual frames as goals, rather than a single goal image configuration.

\paragraph{Efficiency Analysis.} We evaluate the computational efficiency by measuring the average inference time per action step during deployment. Table~\ref{tab:infer_time} presents the average inference latency per step across different methods, measured over 40,000 steps on a single NVIDIA RTX 3090 GPU. Our proposed \method{}(G) demonstrates superior computational efficiency while maintaining state-of-the-art performance. This efficiency stems from our novel approach of dynamically generating weights for a lightweight target policy. Specifically, \method{}(G) generates a suitable set of policy weights at the beginning of each rollout based on the goal image. These weights remain fixed throughout the execution, eliminating the need for repeated weight generation and thus significantly reducing the computational overhead during deployment.

\begin{table}[h]
\centering
\caption{\textbf{Average Inference Time Per Step}}
\begin{tabular}{l|ccc|cc}
\toprule
Method & GCBC & Play-LMP & C-BeT  & \method{} & \method{}(G) \\
\midrule
Time (ms) & 15.47 & 22.78 & 13.61 & 6.33 & 1.46 \\
\bottomrule
\end{tabular}
\label{tab:infer_time}
\end{table}

\section{Comparison with Other Hypernetworks}\label{apx:hpzero}

This section provides additional details on the comparison between our proposed hypernetwork architecture and HyperZero, particularly regarding computational requirements.

\subsection{Training Time \& Memory Requirements}
To ensure a fair and direct comparison, all experiments were conducted on a single NVIDIA RTX 4090 GPU. The training hyperparameters, including batch size, learning rate, and optimizer settings, were kept identical for both our method and the HyperZero baseline. The only modification was the hypernetwork architecture itself. This controlled setup ensures that any observed differences in resource consumption are directly attributable to the design of the hypernetwork module.

As detailed in Table~\ref{tab:resource_comparison}, our method requires slightly more resources than HyperZero in terms of per-epoch training time and memory usage. However, this modest computational overhead is coupled with the substantial performance gains documented in the main paper, highlighting the efficiency and effectiveness of our architectural design.

\begin{table}[h!]
\centering
\caption{
    \textbf{Computational resource comparison.} The table shows per-epoch training time and memory footprint for our method versus HyperZero under identical hyperparameter settings. ``Frozen'' and ``Unfrozen'' refer to the state of the visual encoder.
}
\label{tab:resource_comparison}
\begin{small}
\begin{tabular}{l|cc}
\toprule
\textbf{Metric} & \textbf{HyperZero} & \textbf{Ours} \\
\midrule
Training Time / Epoch    & $\sim$90s   & $\sim$104s   \\
Memory (Frozen Encoder)  & 3,038 MB & 4,916 MB  \\
Memory (Unfrozen Encoder)& 13,452 MB & 14,844 MB \\
\bottomrule
\end{tabular}
\end{small}
\end{table}

\section{Policy Learning Details}\label{apx:learning_details}

\paragraph{Overview.}
A conventional sequential decision-making problem can be formalized as a discrete-time finite Markov decision process (MDP) defined by a 7-tuple $M = (\mathcal{O}, \mathcal{A}, \mathcal{P}, r, \rho_0, \gamma, H)$, where:
\begin{itemize}
\item $\mathcal{O}$ denotes the observation space,
\item $\mathcal{A}$ represents the action space,
\item $\mathcal{P}: \mathcal{O} \times \mathcal{A} \times \mathcal{O} \rightarrow \mathbb{R}_{+}$ defines the transition probability distribution,
\item $\gamma \in [0, 1]$ is the discount factor,
\item $H$ specifies the temporal horizon of the process.
\end{itemize}

In the context of imitation learning, we define a complete state-action trajectory as $\tau = (o_0, a_0, ..., o_t, a_t)$, where the initial state is sampled as $o_0 \sim \rho_0(o_0)$, actions are generated by the policy $a_t \sim \pi_\theta(\cdot|o_t)$, and state transitions follow $o_{t+1} \sim \mathcal{P}(\cdot|o_t, a_t)$.

Traditionally, the objective in goal-conditioned decision-making problems is to identify an optimal goal-conditioned policy $\pi_\theta$ that maximizes the expected discounted reward:

\begin{equation}
\eta(\pi_\theta) = \mathbb{E}_\tau[\sum_{t=0}^T \gamma^t r(o_t, a_t, o_{t+1}|o_g)]
\end{equation}

However, our approach diverges from this conventional framework in several key aspects: \textit{Reward-Free Learning:} Operating within a behavior cloning paradigm, we lack access to explicit reward signals. Instead, we aim to learn a goal-specific policy $\pi_\theta$ that maps states to optimal actions purely from demonstrations.
\textit{Goal-Specific Policy Generation:} Rather than learning a universal goal-conditioned policy, our hypernetwork architecture generates specialized policies for specific goals, conditioned on the current RGB observation and a target goal image.
\textit{Non-Markovian Extension:} We relax the Markovian assumption to incorporate temporal dependencies. The resulting policy formulation becomes:
\begin{equation}
\pi_\theta(a_t|o_{t-1}, o_t)
\end{equation}

This extended formulation enables the policy to leverage information from a context window of length 2, enhancing its capacity to handle complex, temporally-dependent manipulation sequences.

\paragraph{Model Architecture.}
Our architectural design addresses the challenges of visuomotor manipulation tasks, which require processing of high-dimensional visual inputs rather than simple state-based representations. The architecture comprises several key components integrated to handle visual and proprioceptive information effectively. \textit{Visual Processing Pipeline:}
To bridge the gap between high-dimensional visual inputs and hypernetwork processing capabilities, we employ a pre-trained visual encoder~\cite{nair2022r3m} to compress RGB images into compact latent representations. This encoder's training follows a two-phase strategy: {\it Initial phase (first 20 epochs):} Parameters remain frozen to establish stable feature representations; {\it Fine-tuning phase:} Parameters become trainable to optimize task-specific visual features.

\paragraph{Hypernetwork Configuration.}
Our hypernetwork uses the HyPoGen architecture~\cite{renhypogen} with 8 optimization blocks. It processes encoded current and goal images to generate the parameters for a 3-layer MLP target policy, an architecture that can be flexibly extended in depth and width.
This meta-learning approach enables dynamic policy adaptation based on specified goals while maintaining computational efficiency.
Beside the image encoder mentioned above, the action generation incorporates several specialized components:
\textit{Predictive Model:} Implemented as an MLP operating in the compressed latent space, leveraging the reduced dimensionality for efficient dynamics modeling,
\textit{Proprioceptive Encoder:} A compact MLP processes low-dimensional proprioceptive states, providing essential agent state information,
\textit{Feature Integration:} Temporal image features are concatenated with proprioceptive information at each timestep,
\textit{Target Policy:} A lightweight MLP processes the integrated features to generate appropriate control actions.
This architecture efficiently handles the complexity of visuomotor tasks while maintaining computational tractability through dimensionality reduction and feature integration. 

\paragraph{Training Details.} Our framework implements an end-to-end training paradigm with controlled experimental conditions for reproducibility. Using a fixed random seed, we partition the dataset into 950 training and 50 validation demonstrations across all tasks. Training employs a batch size of 256 and the Adam optimizer with an initial learning rate of $5 \times 10^{-4}$, coupled with cosine annealing for learning rate decay. The model trains for 500 epochs without weight decay or dropout regularization in the hypernetwork component. The training and evaluation procedures were performed on a single NVIDIA GeForce RTX 3090 or RTX 4090 GPU. To ensure a fair comparison, all methods evaluated in our experiments were trained using this identical configuration.

\begin{figure}[t]
    \centering
    \includegraphics[width=0.96\linewidth]{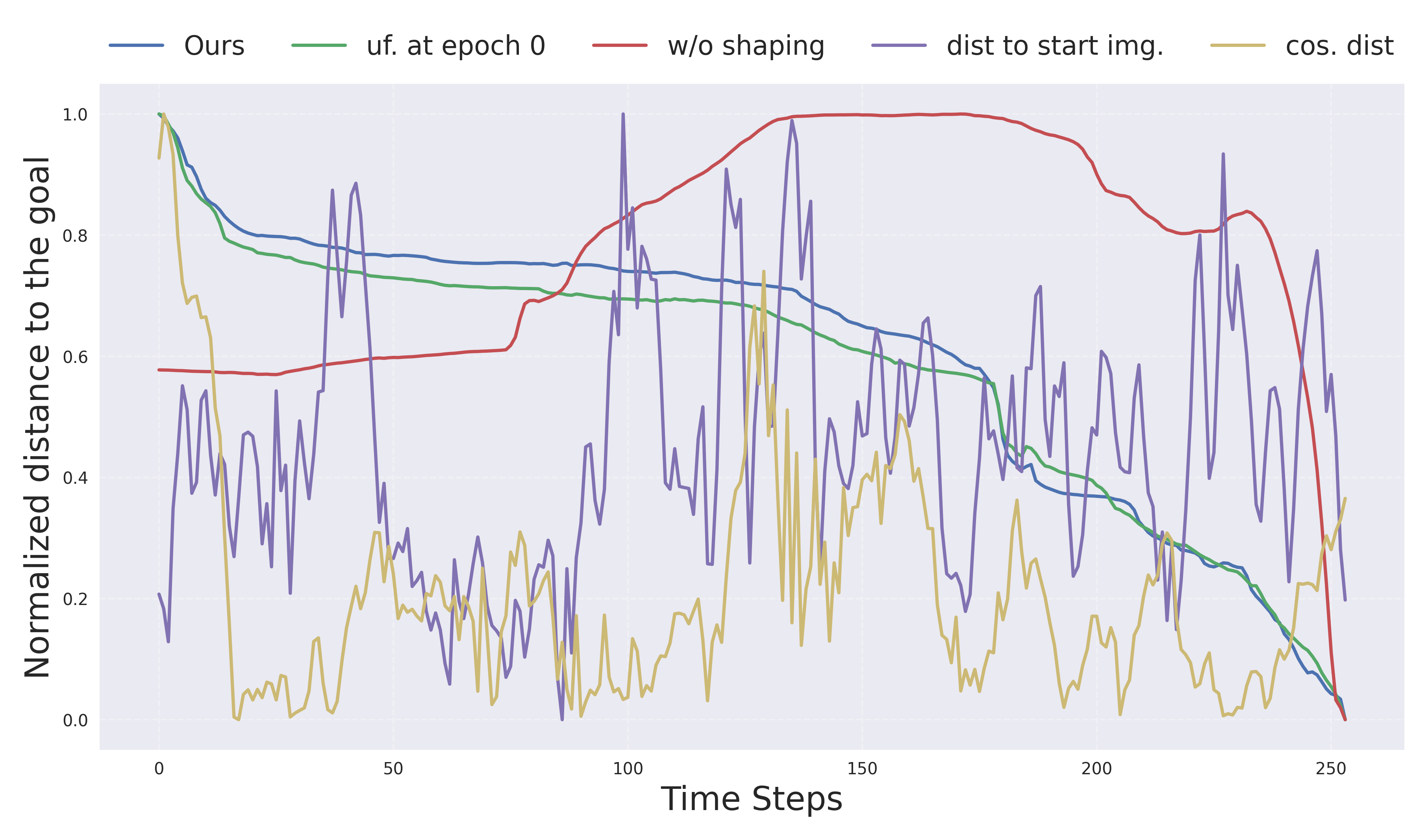}
    \caption{Normalized distance 
    between current and goal states 
    computed with \textbf{different latent spaces} 
    along policy rollouts.}
    \label{fig:normalized_dist}
\end{figure}
\begin{figure*}[t]
  \centering
   \includegraphics[width=0.96\linewidth]{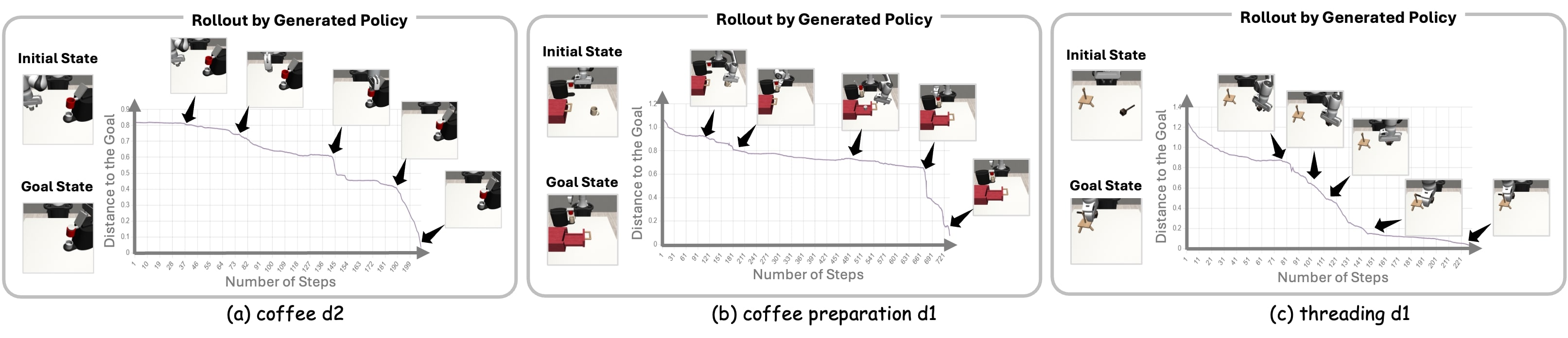}
   \caption{\textbf{
   Distance to the goal in the latent space along policy rollouts across different manipulation tasks.} 
   The consistent decreasing trend across diverse tasks demonstrates that our learned latent representations effectively capture the physical progress toward task goals, establishing meaningful correspondence between latent-space distances and real-world task completion.}
   \label{fig:exp_goalreaching}
\end{figure*}

\section{Ablation Studies}
We conducted ablation experiments to evaluate the impact of various methodological choices, with quantitative results presented in Table~\ref{tab:ablation}. Figure~\ref{fig:exp_goalreaching} and Figure~\ref{fig:normalized_dist} illustrates the normalized distance between current and goal states under different experimental configurations. Our analysis includes several key variations:
\begin{itemize}
    \item uf. at epoch 0. Full parameter unfreezing from epoch 0, where all model components are trainable from initialization.
    \item w/o shapping. Removal of latent shaping technique detailed in Section~\ref{subsec:gr_policy}.
    \item dist$\leftrightarrow$start img. Alternative distance computation between current and start images, rather than current and goal images.
    \item cos. dist. Implementation of cosine distance metric in place of Euclidean distance.
    \item C-Bet(w/ shaping). We implement C-Bet and incorporate our proposed additional shaping method while maintaining the same experimental setup.
\end{itemize}
These systematic variations enable us to quantify the contribution of each design choice to the overall system performance.

\begin{figure}[!t]
    \centering
    \includegraphics[width=0.96\linewidth]{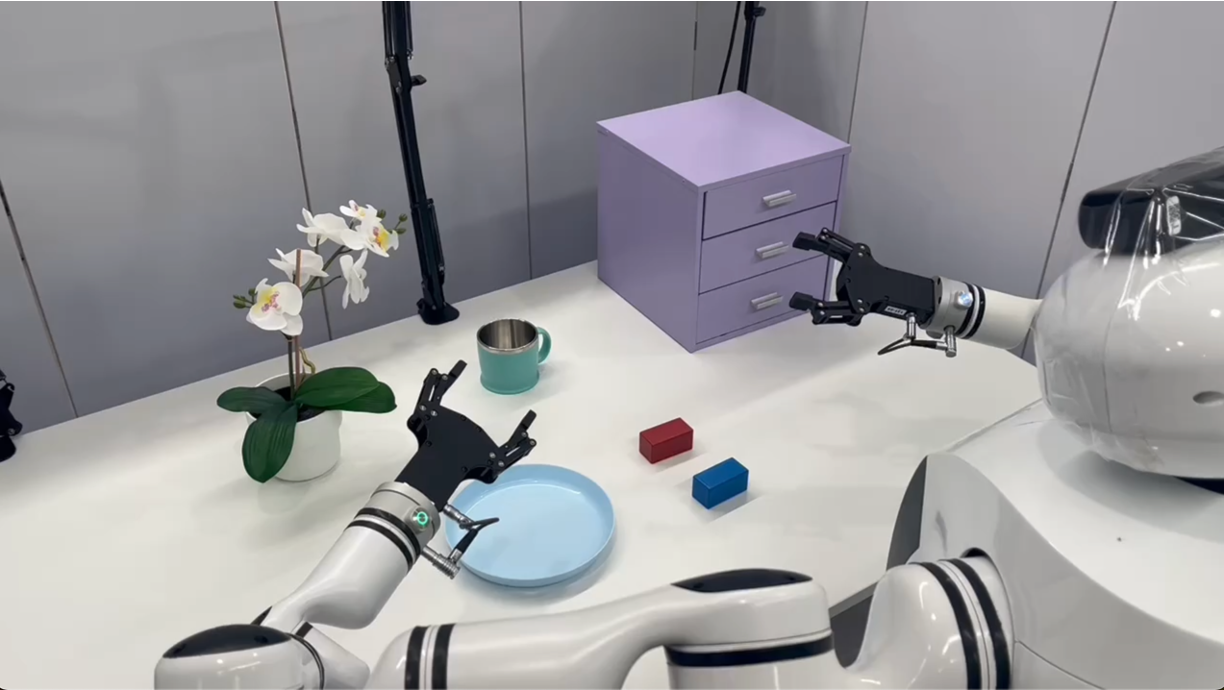}
    \caption{The real robot workspace.}
    \label{fig:workspace}
\end{figure}

\section{Real Robot Experiment Setting}\label{apx:real_setting}
\paragraph{Real Robot Platform.}
All real‐world experiments were conducted on a RealMan RMC‐DA dual‐arm manipulator, which comprises two 7‐degree‐of‐freedom robotic arms, each rated for a 5 kg payload and fitted with a parallel‐jaw gripper. While the platform supports bimanual manipulation, this work focuses specifically on evaluating Hyper-GoalNet's capabilities in single-arm tabletop manipulation tasks. The extension to bimanual manipulation remains as future work. The robot is mounted in front of a 1.25 m × 0.75 m tabletop, which serves as the exclusive workspace for all manipulation tasks. A standardized set of test objects—ranging from simple geometric primitives (e.g., cubes) to more complex shapes—is placed on the table according to predefined configurations. And on the tabletop there might be some distractor object. To support perception, we employ an overhead RGB-D sensor (Intel RealSense D435i). The entire workspace is shown in Figure~\ref{fig:workspace}

\paragraph{Task Details.}
To validate the versatility and robustness of our method across a broad spectrum of manipulation skills, we selected four representative tabletop tasks. Each task emphasizes a different core competency—object localization, precision grasping, and surface contact manipulation—and is defined as follows:

\begin{itemize}
  \item \textbf{Pick-and-Place.}  
    The robot must perceive and localize a specified cubic object within the workspace, plan a collision‐free trajectory, execute a stable grasp, and transport the object to a predefined target location (e.g., a plate). Success is measured by the accuracy of the final placement and the repeatability across trials.

  \item \textbf{Stacking.}  
    Extending the pick‐and‐place paradigm, this task requires the robot to grasp a source cube, position it directly above a target cube resting on the tabletop, lower it until gentle contact is detected via proximity or vision‐based cues, and then release to complete the stack. Success is defined by the source cube being neatly aligned atop the target cube.

  \item \textbf{Drawer Pulling.}  
    The robot must detect drawer handle, plan an approach to engage the handle with its gripper, and execute a controlled pulling maneuver to extend the drawer along its linear guide. Performance is evaluated by the final extension distance achieved without stalling.

  \item \textbf{Sweeping.}  
   The end‐effector is equipped with a broom attachment. The robot must locate and gather a target object scattered on the tabletop, then sweep it into a designated collection zone (e.g., a dustpan). Success is defined by the target object being fully contained within the collection zone at the end of each trial.
\end{itemize}

\begin{figure}[!t]
    \centering
    \includegraphics[width=\linewidth]{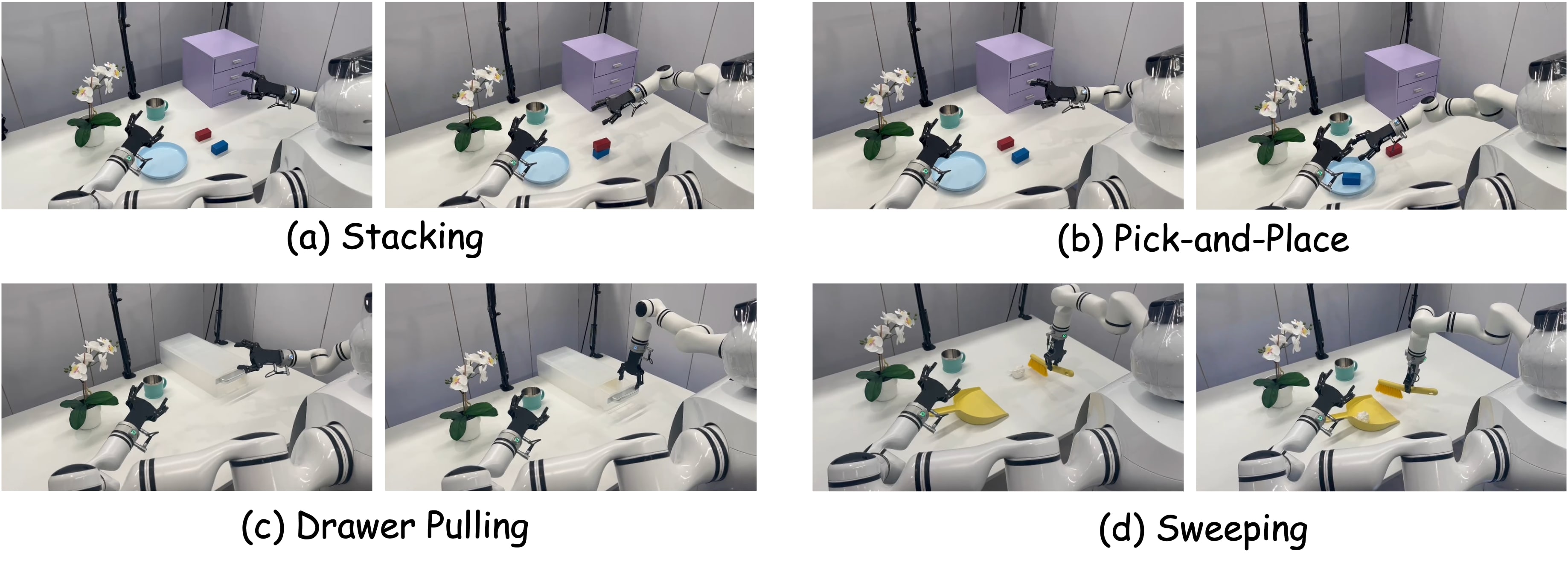}
    \caption{Key phases of diverse manipulation tasks in real robot experimental evaluation.}
    \label{fig:real_tasks}
\end{figure}

\paragraph{Data Processing and Observation Space.}
Our real‑robot evaluation follows the same protocol as in simulation: at each timestep, the policy receives the two most recent observations and a single goal image. We collect approximately 70–100 human teleoperation trajectories per task for training. Vision is acquired with an Intel RealSense D435i depth camera; we concatenate its depth channel with the RGB channels to form 4‑channel images of resolution $128\times128$, which are normalized to $[0,1]$ before input to the encoder. Since explicit end‑effector poses are unavailable, we represent the current proprioceptive state by the previous action—comprising seven joint angles and one gripper command—resulting in an 8‑dimensional vector. All modalities are synchronized identically to the simulation setting, ensuring a seamless transfer between simulated and real‑world experiments.  

\begin{figure*}[!h]
  \centering
  \includegraphics[width=\linewidth]{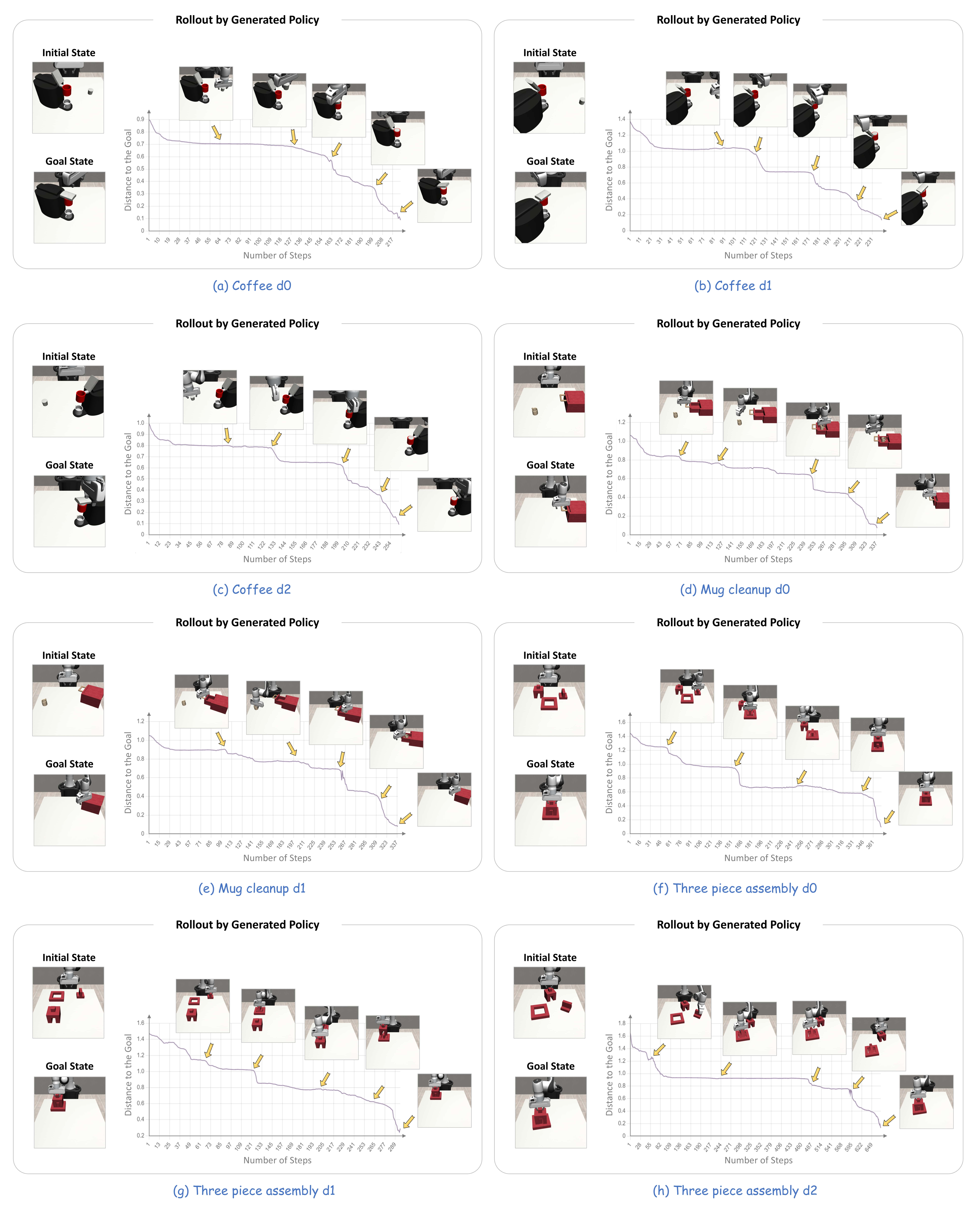}
   \caption{\textbf{
   Additional visualization: Distance to the goal in the latent space along policy rollouts across different manipulation tasks.} 
   The consistent decreasing trend across diverse tasks demonstrates that our learned latent representations effectively capture the physical progress toward task goals, establishing meaningful correspondence between latent-space distances and real-world task completion.}
   \label{fig:more_vis_1}
\end{figure*}

\begin{figure*}[!h]
  \centering
  \includegraphics[width=\linewidth]{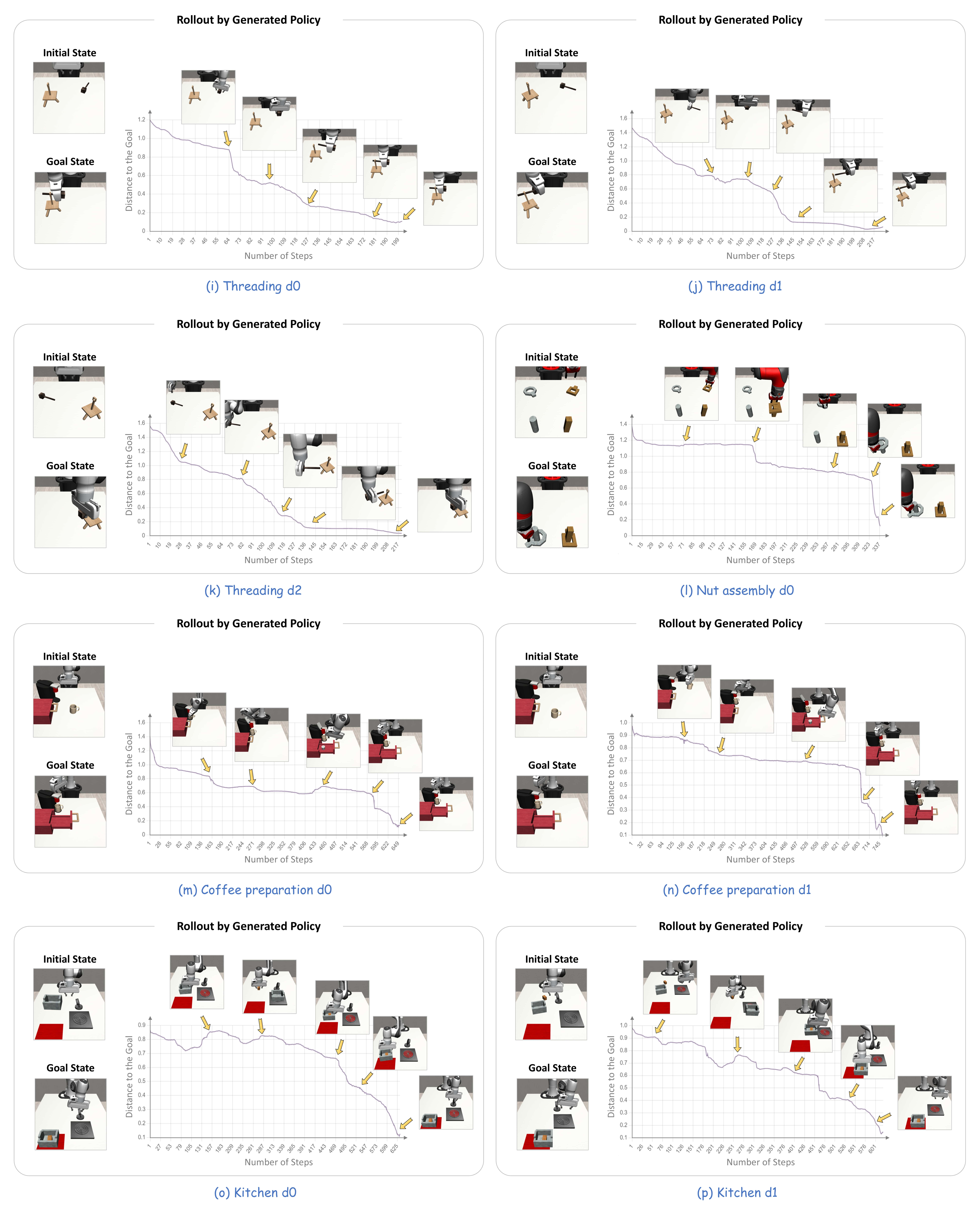}
   \caption{\textbf{
   Additional visualization: Distance to the goal in the latent space along policy rollouts across different manipulation tasks.} 
   The consistent decreasing trend across diverse tasks demonstrates that our learned latent representations effectively capture the physical progress toward task goals, establishing meaningful correspondence between latent-space distances and real-world task completion.}
   \label{fig:more_vis_2}
\end{figure*}

\section{Additional Visualization}

We provide additional visualization of the shaping. Figure~\ref{fig:more_vis_1} and Figure~\ref{fig:more_vis_2} demonstrate the approximate monotonic trend of the distance between the current state and the goal state is consistent across different task scenarios.
This consistent pattern substantiates the robustness of our shaping mechanism and validates its task-agnostic applicability.

\section{Future Work}
Building upon our current findings, we identify three primary directions for future research: First, we aim to incorporate multi-goal reasoning to handle complex sequential tasks. Second, we plan to extend our framework by integrating foundation models to develop a more generalized policy generation framework, potentially enabling broader task generalization and enhanced adaptability across diverse manipulation scenarios. Third, we plan to combine our parameter-adaptive approach with reinforcement learning to reduce dependence on demonstration data.

\end{appendix}

\clearpage

\end{document}